\pdfoutput=1

\documentclass[11pt]{article}

\usepackage{authblk}

\usepackage[]{EMNLP2023}

\usepackage{times}
\usepackage{latexsym}

\usepackage[T1]{fontenc}

\usepackage[utf8]{inputenc}

\usepackage{microtype}

\usepackage{inconsolata}

\usepackage{xspace}

\usepackage{amsmath}
\usepackage{amssymb}

\usepackage{svg}
\usepackage{graphicx}
\graphicspath{ {./images/} }

\usepackage{multirow}
\usepackage{caption}
\usepackage{subcaption}
\usepackage[normalem]{ulem}
\usepackage{booktabs}
\usepackage{amssymb}
\usepackage{pifont}
\newcommand{\cmark}{\ding{51}}%
\newcommand{\xmark}{\ding{55}}%
\usepackage{xcolor}

\definecolor{deepskyblue}{rgb}{0.0, 0.75, 1.0}
\definecolor{lightcornflowerblue}{rgb}{0.6, 0.81, 0.93}
\definecolor{LightCyan}{rgb}{0.88,1,1}
\definecolor{lightpink}{rgb}{1.0, 0.71, 0.76}
\definecolor{pastelpink}{rgb}{1.0, 0.82, 0.86}
\definecolor{teagreen}{rgb}{0.82, 0.94, 0.75}
\definecolor{lemonchiffon}{rgb}{1.0, 0.98, 0.8}
\definecolor{deeppink}{rgb}{1.0, 0.08, 0.58}
\definecolor{deepsaffron}{rgb}{1.0, 0.6, 0.2}
\definecolor{darkpastelgreen}{rgb}{0.01, 0.75, 0.24}
\definecolor{darkorange}{rgb}{1.0, 0.55, 0.0}

\definecolor{overview}{rgb}{0.19, 0.55, 0.91}
\definecolor{research}{rgb}{0.0, 0.29, 0.29}
\definecolor{territ}{rgb}{1.0, 0.75, 0.0}
\definecolor{importance}{rgb}{0.84, 0.04, 0.33}
\usepackage {soul}

\newcommand*\samethanks[1][\value{footnote}]{\footnotemark[#1]}

\title{\textsc{OpenAsp}: A Benchmark for Multi-document Open Aspect-based Summarization}

\author[1\thanks{~~Equal contribution.}]{\bf Shmuel Amar}
\author[1\samethanks~~\thanks{~~Part of the research was conducted during an internship at One AI.}]{\bf Liat Schiff}
\author[1]{\bf Ori Ernst}
\author[2]{\bf Asi Shefer}
\author[3\thanks{~~Work done in cooperation with Bar-Ilan University (external and not related to the author’s work at Amazon).}]{\bf Ori Shapira}
\author[1]{\bf Ido Dagan}
{
\makeatletter
\renewcommand\AB@affilsepx{~~~~~~ \protect\Affilfont} \makeatother

\affil[1]{Bar-Ilan University}
\affil[2]{One AI}
\affil[3]{Amazon}
}
\affil[  ]{} 
\affil[  ]{\tt \{shmulikamar, liatschiff1, oriern\}@gmail.com}
\affil[  ]{\tt asi@oneai.com ~ orishap@amazon.com ~ dagan@cs.biu.ac.il}

\begin{document}
\maketitle

\newcommand{\openasp}{\textsc{OpenAsp}\xspace}

\begin{abstract}
The performance of automatic summarization models has improved dramatically in recent years.
Yet, there is still a gap in meeting specific information needs of users in real-world scenarios, particularly when a targeted summary is sought, such as in the useful aspect-based summarization setting targeted in this paper.
Previous datasets and studies for this setting have predominantly concentrated on a limited set of pre-defined aspects, focused solely on single document inputs, or relied on synthetic data.
To advance research on more realistic scenarios, we introduce \openasp{}, a benchmark for multi-document \textit{open} aspect-based summarization.
This benchmark is created using a novel and cost-effective annotation protocol, by which an open aspect dataset is derived from existing generic multi-document summarization datasets.
We analyze the properties of \openasp{} showcasing its high-quality content. Further, we show that the realistic open-aspect setting realized in \openasp{} poses a challenge for current state-of-the-art summarization models, as well as for large language models.

\end{abstract}
\section{Introduction}


\begin{figure}[ht]
    \centering
    \includegraphics[width=\linewidth]{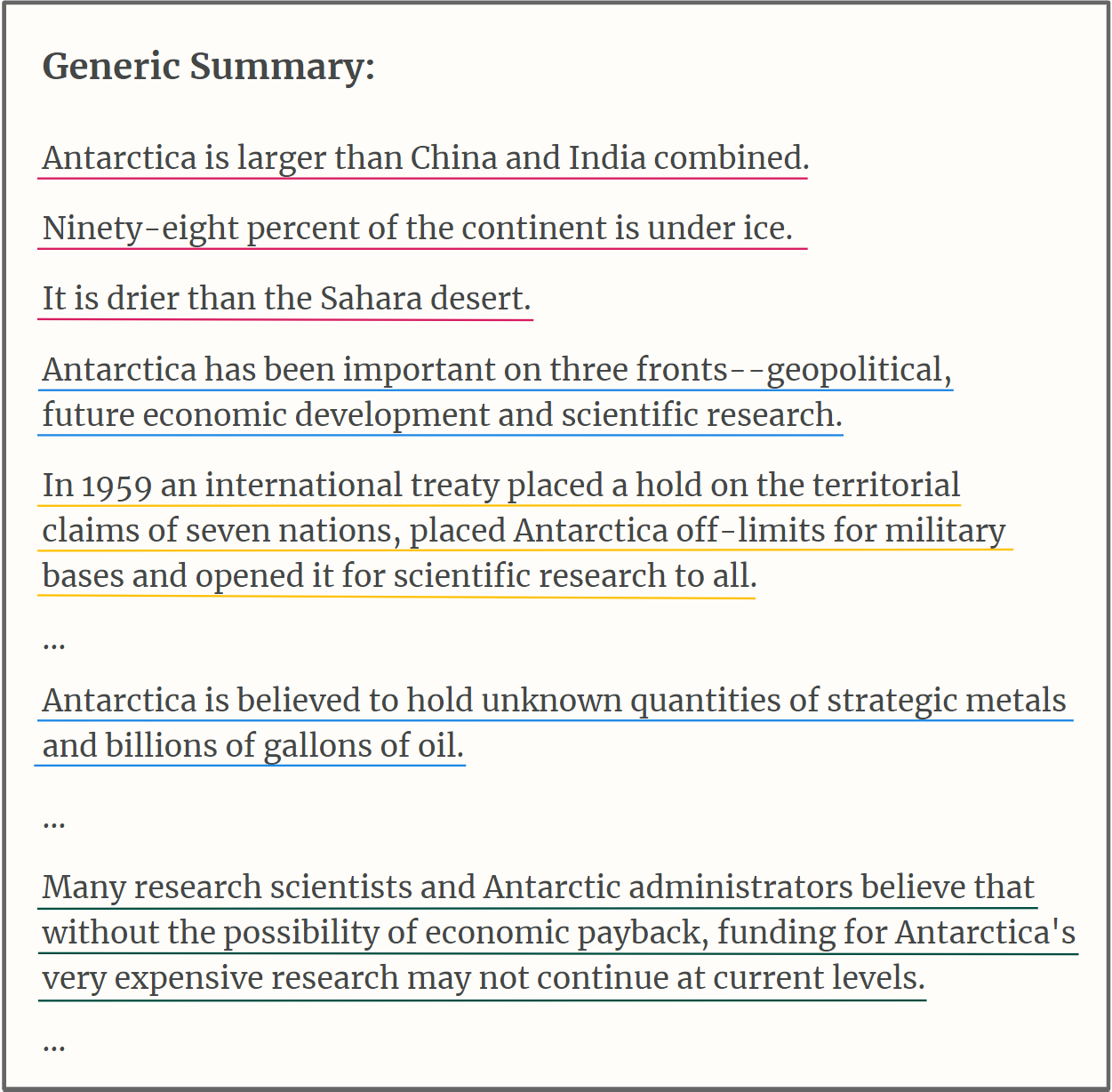}
    \caption{A generic summary from DUC on the topic \textbf{\textit{Antarctica}} with some of the marked sentences for each of the 4 identified aspect labels: 
    \textcolor{overview}{\textit{Overview of Antarctica}}, \textcolor{research}{\textit{Research in Antarctica}}, \textcolor{importance}{\textit{Antarctica's importance}}, and \textcolor{territ}{\textit{Territorial claims}}. The concatenated sentences of an aspect construct the respective aspect-based summary.}
    \label{fig:annotation_ui_main}
\end{figure}


\begin{table*}[t]
\resizebox{\linewidth}{!}{%
\begin{tabular}{cllcclrr}
\toprule
                & \textbf{Dataset} & \textbf{Domain} & \textbf{MDS} & \textbf{Open} & \textbf{Collection} & \textbf{\# Queries$|$Aspects} & \textbf{\# Instances} \\ \midrule
\multirow{4}{*}{\rotatebox[origin=c]{90}{QFS}} & SQuALITY {\small \citep{wang2022squality}}         & Scifi           & \xmark            & \cmark             & manual           & 437                     & 625                   \\
                              & QMSum {\small \citep{zhong-etal-2021-qmsum}}            & Meeting         & \xmark            & \cmark             & manual           & 1,566                   & 1,808                 \\
                              & AQuaMuSe {\small \citep{kulkarni2020aquamuse}}       & General         & \cmark            & \cmark             & automatic        & 5,519                   & 7,168                 \\
                              & TD-QFS {\small \citep{Baumel_Cohen_Elhadad_2016}}           & Medical         & \cmark            & \cmark             & manual           & 40                       & 120                     \\ \midrule
\multirow{5}{*}{\rotatebox[origin=c]{90}{ABS}} & FacetSum {\small \citep{meng-etal-2021-facetsum}}         & Science         & \xmark            & \xmark             & automatic        & 4                       & 60,532                \\

                              & MA-News {\small \citep{frermann-klementiev-2019-inducing}}  & News            & \xmark            & \xmark             & automatic        & 6                       & 286,701               \\
                              & AspectNews {\small \citep{ahuja2022aspectnews}}       & News            & \xmark            & \xmark             & manual           & 4                       & 400                   \\ 
                              & SPACE {\small \citep{angelidis-etal-2021-extractive}}            & Reviews         & \cmark            & \xmark             & manual           & 6                       & 900                 \\
                              & WikiAsp {\small \citep{hayashi-etal-2021-wikiasp}}          & Wikipedia       & \cmark            & \xmark             & automatic        & 200                     & 504,546               \\ \cmidrule{2-8}
                              \multirow{3}{*}{\rotatebox[origin=c]{90}{OABS}} & AnyAspect {\small \citep{tan2020summarizing}}            & News       & \xmark            & \cmark             & automatic        & 363,440               & 2,204,270             \\
                              & OASUM {\small \citep{yang2022oasum}}            & Wikipedia       & \xmark            & \cmark             & automatic        & 1,045,895               & 3,747,569             \\ 
                              & \textbf{\openasp{} (Ours)} & \textbf{News}   & \textbf{\cmark}   & \textbf{\cmark}    & \textbf{manual}  & \textbf{1,266}           & \textbf{1,310}       \\ \bottomrule
\end{tabular}%
}
\caption{Prominent datasets for query-focused summarization (QFS) and abstract-based summarization (ABS). Our annotation protocol enables efficient collection of high-quality multi-document open-aspect-based summaries.
``\# Queries$|$Aspects'' is the number of unique queries or aspects appearing in the dataset, and ``\# Instances'' is the number of data instances in the dataset (document set + query$|$aspect + summary).}
\label{tab:related-datasets}
\end{table*}

When faced with a large body of text, a summary is an effective means to get a concise version of the salient content. However, informational needs of users vary, calling for summarizers that can focus a summary around a given request. The summarization community has addressed this demand mainly through query-focused summarization (QFS) and aspect-based summarization (ABS). Accordingly, several datasets and benchmarks have been compiled over time to enable research on these tasks (see \autoref{tab:related-datasets}). 

In QFS, a query is highly flexible and can target specific information within particular text. In contrast, ABS datasets traditionally predefined small sets of generic subtopics within a common topical category on which aspect-based summaries are generated, such as \textit{geography} and \textit{recovery} aspects in any article about an earthquake \cite{amplayo2021aspect}. \textit{Open}-ABS \citep[OABS;][]{tan2020summarizing}, allows aspects to differ for each source text, yet still just as subtopics in the text.


Collecting datasets for these tasks is a major obstacle. This kind of data does not naturally occur in available resources, and manually annotating is highly burdening, particularly for the multi-document setting. Indeed, the very few existing OABS datasets are synthetically gathered. In addition, they address a single-document setting, even though real-world scenarios involve information navigation within \textit{multiple} documents on a topic.

In this work, we first propose a novel efficient protocol to \textit{manually} derive high-quality multi-document open aspect-based summaries from standard multi-document summarization (MDS) datasets. Through crowdsourcing, open aspects and their summaries are extracted from generic reference summaries. 

Applying our protocol, we present \openasp{}, an OABS dataset for the multi-document setting.\footnote{The \openasp{} dataset is available at \url{https://github.com/liatschiff/OpenAsp}.} The aspects and respective summaries are based on the prominent DUC \citep{nist2002DUCWebsite} and MultiNews \citep{fabbri-etal-2019-multi} datasets. See \autoref{fig:annotation_ui_main} for an example of open-aspect-based summaries extracted from a generic summary. 
The dataset contains 1,310 aspect-based summaries, split to train, validation and test sets, enabling methodological modeling for the task.
We further implement and analyze several baseline models that demonstrate the challenging nature of the task and dataset, even for recent high-performing large language models.

\section{Background}
\label{sec:background}

QFS is a long-standing task that addresses the need to summarize around a specified user request \citep{dang2005overview}. A query is inherently fluid, allowing great variance in length, specificity and format \citep{carmel2006query}. Corresponding datasets, such as those at the top of \autoref{tab:related-datasets}, mainly evolved around queries for specific information needs within the source text(s). \citet{wang2022squality}, for instance, focused summaries around questions targeting distinct matters within the document, e.g. summarizing around ``What is the CPA and what does it do?'' in a certain story.

Early research on ABS \cite{Hu2004MiningAS, titov-mcdonald-2008-joint} recognized the need for more structured information around subtopics of the source text. Relevant datasets (middle of \autoref{tab:related-datasets}) focus on recurring aspects within a domain, and approach this by pre-defining a fixed list of aspects, e.g., ``service'' and ``price'' in all restaurant reviews.
Some datasets expanded to multi-document inputs \cite{angelidis-etal-2021-extractive,hayashi-etal-2021-wikiasp}
which is a more realistic setting when seeking topical information.

An additional direction stemming from ABS permits \textit{open} aspects that still concisely target subtopics, but can be unique for the individual input text (bottom of \autoref{tab:related-datasets}, with more details in \autoref{tab:open-datasets} in the appendix). 
The existing OABS datasets, namely AnyAspect \cite{tan2020summarizing} and OASUM \cite{yang2022oasum}, are synthetically compiled and only address single document inputs.
In AnyAspect, the named entities within the document are considered the aspects, and the corresponding summaries include any source sentence mentioning the respective entity.
In OASUM, the aspects and the summaries were extracted from Wikipedia articles, where Wikipedia sub-titles are the aspects and their summaries are automatically extracted from the article’s abstract section via a greedy lexical similarity method.
Applying such artificial methods yield lower-quality input documents, aspects and expected output summaries.

Collecting datasets for the ABS task poses a challenging undertaking that includes reading large input texts and writing out the focused summaries. Unlike in generic summarization, where large sources of summaries can be scraped from the web \cite{hermann2015teaching, grusky-etal-2018-newsroom, fabbri-etal-2019-multi}, summaries for our setting are not generally available. 
Meanwhile, manually writing high quality aspect-based summaries from scratch is an expensive labor-intensive task. \citet{wang2022squality} reported 20--40 minutes just for reading a 3000--6000 word story.
Summarizing multi-document sets is even more complex since the total input-length may be much larger (e.g., tens of thousands of words; see Section \ref{sec:dataset-analysis}), while information-overlap further requires content consolidation by the summarizer. 
For instance, \citet{dang2005overview} reported 5 hours of labor for generating one multi-document summary by an expert.
Employing crowdsource workers, on the other hand, has been shown to lead to poor extractive multi-document summaries \cite{DBLP:journals/lre/LloretPA13}.
Our protocol exploits existing MDS benchmarks and applies controlled crowdsourcing \cite{roit-etal-2020-controlled}, and is hence substantially cheaper and more efficient than previous manual collection processes.

Our \openasp{} dataset addresses all the above issues by supporting open-aspects in the multi-document setting, with manually annotated realistic summaries. Moreover, the annotation protocol can be applied across any available generic summarization dataset to produce even more like-quality aspect-based summaries.
\section{Task Formulation}\label{sec:task-formulation}

Following prior work \cite{ahuja2022aspectnews, yang2022oasum}, given a set of texts about a topic, we define an aspect as a central theme within a topic.
The aspect can be referred by certain phrases,
denoted \emph{aspect labels}. As an example, \textit{Research in Antarctica} and \textit{Territorial claims} are aspect labels of the \textit{Antarctica} topic (see \autoref{fig:annotation_ui_main}).

Similar to previous work on ABS \citep{hayashi-etal-2021-wikiasp, angelidis-etal-2021-extractive}, our aspect label is short and concise. In contrast, our aspect definition is \emph{open} allowing ad-hoc aspects with free-form labels, contrary to having pre-defined domain-specific aspects.
Relative to a \textit{query} in query-focused summarization \citep[QFS;][]{dang2005overview}, which might specify a complex information need, our aspects are restricted to relevant subtopics.
\cite{hayashi-etal-2021-wikiasp,angelidis-etal-2021-extractive,angelidis-lapata-2018-summarizing}.

The OABS task definition follows previous work \cite{tan2020summarizing, yang2022oasum},
and is extended to the multi-document setting as follows:
Given a set of documents $D$ on the same topic and an arbitrary aspect label $a$, the task is to output a short aspect-based summary $S^a$. The summary should consolidate salient information from the document set that is relevant to the aspect.
\section{Annotation Protocol}\label{sec:annotation_protocol}



As emphasized in Section \ref{sec:background}, manually collecting aspect-based summaries is very costly.
We propose a novel and cost-effective protocol for generating aspect-based multi-document summaries, executed through controlled crowdsourcing \citep{roit-etal-2020-controlled} and a specially-designed annotation tool (\autoref{fig:annotation_ui_full} in the Appendix). The key idea of our protocol is the extraction of gold aspect-based summaries from \textit{generic} summaries in existing MDS datasets.
Notably, the process is accomplished by reading the generic summary text only, as described below, while saving the strenuous need to read the entire set of source documents and to write the aspect-based summary from scratch.

\subsection{Collecting Aspects and Summaries}
\label{sec:protocol}
From an existing MDS dataset, we gather pairs consisting of a document set $D$ and a respective generic summary $G$. An annotator reads $G$ and identifies prominent aspects within it, specified by aspect labels $a_{1}, a_{2}, ..., a_{m}$.
For each identified aspect label $a_{i}$, the annotator selects the relevant sentences from $G$.
The concatenation of these sentences, retaining the original sentence-order from $G$, produces the corresponding aspect-based summary $S^{a_i}$.
Accordingly, we establish $m$ new aspect-based summaries for $D$ as instances for the dataset. Notice that a summary is abstractive with respect to $D$, being comprised of sentences from the abstractive generic reference summary.

In our process, we favor extraction of fewer but high quality aspects from a generic summary.
Specifically, our protocol instructs annotators to detect the aspects that are central in the generic summary, and to avoid circumstantial aspects.
Although our protocol does not exhaustively extract aspects for the topic, the main sub-topics found in the generic summary establish a reliable and sufficient sample of aspects for addressing the multi-document open ABS task, for training and evaluating models.
The full annotation guidelines appear in Appendix \ref{appendix:annotation-guidelines}.

Critically, the described protocol avoids reading through the full document set and writing text for the summary. Instead, each aspect summary comprises a subset of generic summary sentences. 
We suggest that summary quality is maintained since the extracted summaries are based on dependable generic gold summary sentences. 
The validity of our protocol is based on two assumptions: (1) the aspect-related sentences extracted from generic summaries cover well the prominent information about the aspect within the full source document-set; (2) the aspect-based summaries preserve the coherence borrowed from the source summaries. 
We show that these assumptions indeed hold by assessing our collected dataset in Section \ref{subsec:dataset-eval}.

\subsection{Curation Phase} \label{sec:curation-phase}
We propose an optional curation phase for cleaning the annotated aspect labels and corresponding summaries. The process encompasses a manual review, by an expert, of the aspect label and aspect-based summary only. The reviewer can edit the aspect label, remove irrelevant sentences from the summary, or completely reject the aspect.
Similar to the annotation protocol, the curation phase avoids the expensive task of reading the source documents.
\section{The \openasp{} Dataset}
\label{sec:dataset}

\subsection{Source Data}
\begin{table}[t]
\centering
\begin{tabular}{l|ccc}
\toprule
Split & \# Topics & \# Instances & \# Docs \\  \midrule
Test & 192 & 596 & 6,536 \\ 
Valid & 82  & 238 & 2,168 \\ 
 Train & 145 & 476 & 4,878 \\
\bottomrule
\end{tabular}
\caption{The size of the \openasp{} dataset splits. ``\# Topics'' denotes the number of document sets, ``\# Instances'' is the total number of aspect-based summaries, and ``\# Docs'' is the total number of source documents.}
\label{tab:dataset-splits-count}
\end{table}

We exploit 2 prominent MDS datasets that contain reference summaries with at least 200 words to demonstrate our protocol robustness: 
DUC,\footnote{\url{duc.nist.gov}; we use DUC 2006-07, and DUC 2001-02 task 2 (that contain 200-word summaries).} a high-quality and expert-annotated dataset, and MultiNews \citep{fabbri-etal-2019-multi}, with data scraped from \url{newser.com}.
For MultiNews, we automatically filtered out samples with invalid source documents, to avoid consequential hallucinations in the summaries (see Appendix \ref{appendix:data-sources}). The large scale of MultiNews allowed further filtration to capture only instances with summaries of 350--880 words, to increase the potential yield of aspect-based summaries. For all source data, we excluded document-set instances that discuss topics presented as a list of related events (e.g., daily news briefs or
various unrelated incidents of the same kind), since 
the generic summaries of such instances typically contain few subtopics, if any.

\subsection{Dataset Collection}
\label{sec:dataset-collection}
We followed the annotation protocol described in Section \ref{sec:protocol}. Specifically, we used controlled crowdsourcing \cite{roit-etal-2020-controlled} for selecting 3 annotators on Amazon Mechanical Turk\footnote{\url{https://www.mturk.com/}} that successfully completed an introductory summary annotation task and correctly answered followup questions on the task guidelines.\footnote{Workers were paid \$0.9 and \$0.6 bonus per task with an average task completion time of about 6 minutes, resulting in \$15.00/hr as recommended by \citealp{Whiting_Hugh_Bernstein_2019}.}

Our workers annotated
236
generic summaries from MultiNews and
208
from DUC. From a total of 
444
generic summaries, annotators extracted 
1,455
aspect-based multi-document summaries. We (paper authors) then applied the curation procedure (Section \ref{sec:curation-phase}) on
1,173
aspect based summaries as detailed in Appendix \ref{app:dataset-curation-details}.\footnote{Staring from the test and validation sets, and moving on to the train sets. We eventually excluded the MultiNews train set instances in the curation process as pass rates for the other sets were high enough.}
Out of the reviewed summaries, we modified 
152
aspect labels, edited sentence choice of
48
summaries, and completely rejected 
94
aspect based summaries (92\%
pass rate). Overall, we gathered 
1,361
summaries for \openasp{}, averaging 3 aspect-based summaries per topic (document set instance), and costing $\sim$\$0.5 per summary.

We split \openasp{} into train, validation and test sets, keeping the original MultiNews splits and splitting DUC datasets by years (Appendix \ref{app:dataset-source-splits}). We set aside 51 summaries (from 16 topics) from the test and validation sets, denoted \textit{analysis-test} and \textit{analysis-val} sets, for quality assessment and modeling (Sections \ref{sec:analysis} and \ref{EvaluationAndResults}).
Statistics on the final \openasp{} sizes appear in \autoref{tab:dataset-splits-count}.
\section{\openasp{} Assessment}
\label{sec:analysis}

We next examine the quality of the collected data, and then analyze its properties.

\subsection{Dataset Quality}
\label{subsec:dataset-eval}

We applied a manual evaluation process to verify the collected summaries' expected qualities. Following \citet{fabbri-etal-2021-summeval}, a summary should be measured for 4 quality criteria: (1) \textit{relevance}, the selection of important content from the source; (2) \textit{coherence}, the quality of the collective structure of all sentences; (3) \textit{consistency}, the factual alignment between the summary and the summarized source; and (4) \textit{fluency}, the linguistic quality of individual sentences. 
In the OABS setting, the \textit{aspect-relevance} is an additional expected quality \citep{amplayo2021aspect, angelidis-etal-2021-extractive}.
This criterion inspects whether the summary includes information that is relevant to the paired aspect.

We assess the 5 quality criteria on 20 aspect-based summaries sampled from the analysis-test set (Section \ref{sec:dataset-collection}). A summary was rated on a 1--5 scale for each criterion by one expert and reviewed by another. In case of a disagreement, the two raters resolved the dispute through reconciliation.\footnote{The reviewer-to-rater agreement was 0.753 linear weighted Cohen's Kappa, and the reviewer-to-reconciled agreement was 0.847, indicating ``substantial'' (0.6--0.8) and ``almost perfect'' (0.8-1.0) agreement respectively.}


The relevance and consistency criteria require comparison of the evaluated summary against the aspect-relevant information across the source document-set. Therefore, for each aspect in the analysis sets, we extracted (via crowdsourcing) all the sentences in the corresponding document-set related to the aspect.\footnote{We measured 0.642 Cohen's Kappa for inter-rater agreement of per sentence aspect label, indicating ``substantial'' agreement.}



The average ratings can be found in \autoref{tab:dataset-eval-results}. The high \textbf{relevance} score of 4.6 supports our first extraction protocol assumption (Section \ref{sec:protocol}) that the aspect-based summaries cover the most important information about the subtopic, even though they originate from generic reference summaries.
\textbf{Consistency} is expectedly sturdy as well, since sentences are copied from gold generic summaries. Hence, consistency issues that are not already present in the generic summaries should not be introduced.
Similarly, the \textbf{fluency} of sentences is adopted from that of the source reference summaries, which is almost flawless.

Although summaries that are extracted from another text can easily suffer from \textbf{coherence} issues, this is rarely the case with our protocol.
We found that 86\% of consecutive sentence pairs in our aspect-based summaries are respectively consecutive in the generic summary. Specifically, 60\% of the aspect-based summaries are full continuous sentence sets from the generic summary. This phenomenon occurred naturally during annotation, without explicit instructions to follow such a principle. Consequently, coherence is also maintained according to the source generic summaries. Overall, coherence scored very high as well, validating our second assumption that our protocol generates coherent aspect-based summaries even as they are based on generic summaries.

To empirically corroborate the quality of the aspects, we statistically analyzed the source sentences corresponding to each of the 51 aspects in our analysis sets. We found that, on average, an aspect relates to 13.5\% of all source documents' sentences and appears in 68\% of the source documents, indicating its topical dominance. Furthermore, only 11\% of all aspect-related sentences refer to more than one aspect, indicating the high level of distinctness of the aspects. Finally, on average, 40\% of all source sentences in a document set (topic) are related to at least one of the aspects extracted for that topic, indicating the substantial coverage of the set of aspects with respect to the entire document set content. Taken together, these findings establish that the collected aspects are central to the topic, covered thoroughly by its documents and collectively cover a main portion of its content.

Overall, the aspect-based summaries representing \openasp{} were determined to exhibit high reliability for the OABS task, consistent with that of the standard generic MDS datasets from which they were extracted.

\begin{table}[t]
\resizebox{\columnwidth}{!}{
\begin{tabular}{ccccc}
\hline
\textbf{Coherence}   & \textbf{Consistency} & \textbf{Fluency}     & \textbf{Relevance}   & \textbf{Aspect Rel.} \\ 
4.7 (0.92) & 4.8 (0.55) & 5.0 (0.22) & 4.6 (0.68) & 4.7 (0.57)      \\ \hline
\end{tabular}}
\caption{Average (std) human evaluation ratings (1--5 scale) on the five quality criteria, determined for 20 instances from the analysis-test set.}
\label{tab:dataset-eval-results}
\end{table}

\subsection{Dataset Analysis}
\label{sec:dataset-analysis}


We discuss properties of \openasp{} that emphasize its underlying diversity from several angles. Details for these analyses are available in Appendix \ref{sec:appendix_dataset_details}.

The input lengths, measured as the aggregated document lengths in a topic, varies from several hundred to tens of thousands of tokens (\autoref{fig:data-lengths-full} in the Appendix), averaging 7,930 words. The summary length ranges from tens to hundreds of tokens, with a median input-to-output (compression) ratio of 69:1.

A topic contains 1--7 aspects with an average of 3.1 aspects per topic (see \autoref{fig:aspect-counts-pie} in the appendix). The aspect labels are almost all lexically unique, repeating on average 1.1 times throughout the dataset. The aspect label is 1--10 words, averaging 3.6 words. Some examples of topics and aspects from \openasp{} appear in \autoref{tab:dataset-detailed-examples}. Since aspect labels in \openasp{} are annotated as sub-topics of their corresponding topic, and are flexibly scripted, the aspects naturally vary widely.

\begin{figure}[t]
    \centering
    \includegraphics[width=0.8\linewidth]{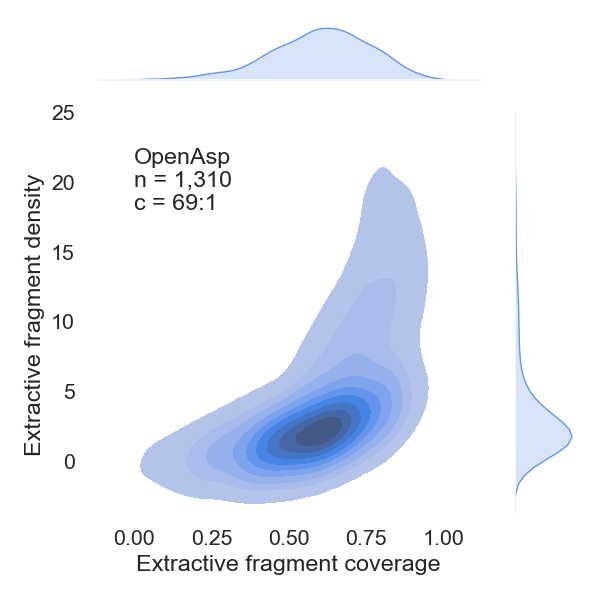}
    \caption{\openasp{} content diversity \cite{grusky-etal-2018-newsroom}. $n$ and $c$ denote number of examples and median compression ratio of \openasp{}, respectively. The large area of the plot indicates that the dataset contains aspect-based summaries with a balance of extractive and abstractive fragments. (See Appendix \ref{sec:appendix_content_diversity} for further explanations.)}
    \label{fig:content_diversity_main}
\end{figure}

Due to our annotation protocol, the summary-abstractiveness in source datasets (DUC and MultiNews) is transferred to the summaries in \openasp{}. Accordingly, the aspect-based summaries exhibit varying extents of abstractivness, as apparent in the diversity plot in \autoref{fig:content_diversity_main} \cite{grusky-etal-2018-newsroom, fabbri-etal-2019-multi}. Consequently, \openasp{} requires models to perform well on both extractive and abstractive forms of summarization.

Finally, a topic consists of 2--25 documents, with an average of 10.4 documents per topic (\autoref{fig:aspect-counts-pie} in the appendix). Following \citet{wolhandler-etal-2022-multi}, we find that the aspect-based summaries rely on a varying number of corresponding source documents (\autoref{fig:multimds-plot} in the appendix). In practice, this means some summaries require handpicking information from specific documents, while others require consolidating information from across the input document set.

Overall, the analyzed properties expressly show the diversity of \openasp{} and the ensuing challenges of the task.

\section{Baseline Models}\label{BaselineModels}
\newcommand{\ChatGPTSumm}{ChatGPT\textsubscript{Summ}\xspace}
\newcommand{\ChatGPTStages}{ChatGPT\textsubscript{Recursive}\xspace}
\newcommand{\ChatGPTStageslong}{ChatGPT-16k\textsubscript{Recursive}\xspace}
\newcommand{\ChatGPTSummBold}{\textbf{ChatGPT\textsubscript{Summ}}\xspace}
\newcommand{\ChatGPTStagesBold}{\textbf{ChatGPT\textsubscript{Recursive}}\xspace}
\newcommand{\ChatGPTStageslongBold}{\textbf{ChatGPT-16k\textsubscript{Recursive}}\xspace}

\newcommand{\BARTStages}{BART\textsubscript{Recursive}\xspace}
\newcommand{\PRIMERAStages}{PRIMERA\textsubscript{Recursive}\xspace}
\newcommand{\BARTStagesBold}{\textbf{BART\textsubscript{Recursive}}\xspace}
\newcommand{\PRIMERAStagesBold}{\textbf{PRIMERA\textsubscript{Recursive}}\xspace}

\newcommand{\BARTSumm}{BART\textsubscript{Summ}\xspace}
\newcommand{\PRIMERASumm}{PRIMERA\textsubscript{Summ}\xspace}
\newcommand{\BARTSummBold}{\textbf{BART\textsubscript{Summ}}\xspace}
\newcommand{\PRIMERASummBold}{\textbf{PRIMERA\textsubscript{Summ}}\xspace}

\newcommand{\BARTopenAsp}{BART\textsubscript{OpenAsp}\xspace}
\newcommand{\BARTopenAspBold}{\textbf{BART\textsubscript{OpenAsp}}\xspace} 
\newcommand{\PRIMERAopenAsp}{PRIMERA\textsubscript{OpenAsp}\xspace}
\newcommand{\PRIMERAopenAspBold}{\textbf{PRIMERA\textsubscript{OpenAsp}}\xspace} 
\newcommand{\OracleSelector}{Oracle\textsubscript{Sel}\xspace}

In this and the subsequent sections we demonstrate the challenges that our dataset lays out for summarization models, and suggest initial directions to cope with these challenges.
A major hurdle to overcome is the large input length of a document set, averaging $\sim$8K tokens in our data, and stretching to $\sim$30K (Section \ref{sec:dataset-analysis}). 
Even with current advancements made to support growing input sizes, properly attending to relevant information in a large input remains a hurdle. There were no feasible models available to us that would fit all of our document-sets.

To cope with this we investigate two common schemas used in the MDS setting:
(a) \textit{Filter-then-summarize} \citep[e.g.,][]{wang2022squality, Baumel_Cohen_Elhadad_2016}, runs a sentence selector to extract aspect-related sentences,
and passes them on to a summarization model (\ref{subsec:decomposed-models});
(b) \textit{Recursive summarization} \citep[e.g.][]{shapira2020massive} summarises subsets of documents separately, and then summarizes these summaries (\ref{subsec:end2end-models}).

\subsection{Filter-then-Summarize}
\label{subsec:decomposed-models}


\paragraph{Sentence selection.}

To slim down the input, one remedy is to select sub-texts that are more likely to be included in a summary.
In MDS in the news domain, the conventional technique for this is the \textbf{Lead} method, which extracts a number of sentences from the beginning of each document, since these often contain the most crucial information in a news report.

However, in the case of ABS, there is no guarantee that salient aspect-specific information appears at the beginning of each document.
Standard ABS models with a closed set of aspects can train a classifier to map a sentence to an aspect \citep[e.g.][]{hayashi-etal-2021-wikiasp}.
In contrast, our open-aspect setting demands a selector that is robust to any aspect. We leverage the \textbf{Sentence-T5} model \cite{ni2021sentence} as an unsupervised sentence selector. Specifically, a cosine similarity score is computed between the dense vectors of an aspect label and a sentence, producing a ranking of aspect-relevant sentences.


For the summarization models we employ in the subsequent stage, we restrict the applied sentence selector to provide a set of sentences that fit within the input-size limit (1K or 4K tokens). An approximately equal number of sentences from each document is used within the limit. 



\paragraph{Summarization.}
For the second stage of the decomposed procedure, we employ three architecturally different summarization models. The first two are sequence-to-sequence models trained for generic summarization: \textbf{BART} \cite{lewis2019bart} was pre-trained on the CNN/DailyMail dataset \citep{hermann2015teaching} for single-document summarization, while \textbf{PRIMERA} \citep{xiao2021primer} was trained on the MultiNews dataset \citep{fabbri-etal-2019-multi} for the multi-document setting. Accordingly, the former has a limit of 1K tokens, and the latter 4K. While PRIMERA is more suitable for our multi-document setting, BART has shown strong performance on single-document ABS \citep{tan2020summarizing, meng-etal-2021-facetsum, wang2022squality} and is a worthy candidate. BART is adjusted for multi-document inputs with separation tokens between documents, following PRIMERA. We further fine-tuned the two models with the \openasp{} train set, denoted \BARTSummBold and \PRIMERASummBold, respectively.\footnote{We concatenate documents (only with the select sentences from the first stage) with a separator, with the aspect at the beginning and end of the input \citep[inspired by][]{wang2022squality}.}

In addition, we experiment with OpenAI's promising
ChatGPT LLM, based on \texttt{gpt-3.5-turbo-0301} \citep{openai2023api}, denoted \ChatGPTSummBold. Such prompt-based models have demonstrated promising results on the zero-shot setting for summarization \cite{zhang2023summit}, which we apply here as well.
See Appendix \ref{appendix:chatgpt-details} for technical details.

\subsection{Recursive Summarization}
\label{subsec:end2end-models}

A different approach to handle the large input size is summarizing subsets of the input that fit the model size limit. The subset summaries are then recursively summarized to generate the final summary (see details in the Appendix \ref{appendix:chatgpt-details}). We experiment with ChatGPT on a 4K or 16K input token-size, denoted \ChatGPTStagesBold, whose robustness enables the recursive scheme without any fine-tuning. About 75\% of our test instances require recursion for the 4K limit, however only about 10\% don't fit within the 16K limit, rendering nearly end-to-end summarization on our test set.

For the sake of completeness, we also experimented with BART and PRIMERA
in the recursive scheme, denoted \BARTStagesBold and \PRIMERAStagesBold, respectively. 
Note that training for this approach would require gold aspect-based summaries for \textit{subsets} of the document set, which are not available.
We therefore activated, in the recursive technique, the BART and PRIMERA systems that were fine-tuned for the filter-then-summarize approach above (Section \ref{subsec:decomposed-models}).

\section{Baseline Evaluation and Analysis}
\label{EvaluationAndResults}



To assess the overall capabilities of our baseline models, we show an overall comparison of the methods (\S{\ref{subsec:eval_automatic}}), an ablation analysis on the \BARTSumm-based configurations (\S{\ref{subsec:eval_analysis}}), and end with a human evaluation on our best baselines (\S{\ref{subsec:eval_human}}).



\begin{table}[t]
\resizebox{\columnwidth}{!}{%
\begin{tabular}{l|ccc|ccc}
\toprule
\multicolumn{1}{c|}{\multirow{2}{*}{\textbf{Sentence Selector}}} & \multicolumn{3}{c|}{1024 Tokens}                                             & \multicolumn{3}{c}{4096 Tokens}                                             \\
\multicolumn{1}{c|}{}                                                    & \multicolumn{1}{c}{R} & \multicolumn{1}{c}{P} & \multicolumn{1}{c|}{$F_1$} & \multicolumn{1}{c}{R} & \multicolumn{1}{c}{P} & \multicolumn{1}{c}{$F_1$} \\ \midrule
Lead                                                                     & 16.8                    & 19.0                    & 17.8                     & 55.9                    & 15.4                    & 24.1                    \\
Sentence-T5                                                              & \textbf{31.9}           & \textbf{37.3}           & \textbf{34.4}            & \textbf{71.8}           & \textbf{21.0}           & \textbf{32.5}           \\ \bottomrule
\end{tabular}
}
\caption{$F_1$ scores, and corresponding Recall (R) and Precision (P) scores, comparing the two sentence selectors against gold sentence alignments from the combined \textit{analysis-valid} and \textit{analysis-test} sets.
In both selectors the top sentences that fit within 1024 or 4096 tokens are extracted, and evaluated separately.}
\label{tab:sentence-selector}
\end{table}

\subsection{Automatic Evaluation}
\label{subsec:eval_automatic}

\paragraph{Sentence selection.}
We first compare the Lead and Sentence-T5 sentence selectors, measuring the relevance of selected sentences to paired aspects. To that end, we utilized the sentence alignments (between aspects and document-set sentences) from the combined \textit{analysis-valid} and \textit{analysis-test} sets (Section \ref{subsec:dataset-eval}), consisting of 1,782 aspect-sentence pairs. For the two sentence selectors, we selected sentences up to a cap of 1K or 4K tokens, and measured the $F_1$ scores between the gold alignment pairs and the resulting selector's pairs, over all aspects. As expected, the results in \autoref{tab:sentence-selector} strongly favor Sentence-T5, which directly focuses on a given aspect label. The lead sentences from the documents are much less relevant to any given aspect.

\paragraph{Summary quality.}
We next assess the aspect-based summaries of our baseline filter-then-summarize and recursive summarizers. We apply the commonly used ROUGE metrics \cite{lin2004rouge}, which measure the lexical overlap between the system and reference aspect-based summaries.
Here we only apply the Sentence-T5 sentence selector in the filter-then-summarize configuration, as it is the better of the two selectors, as observed above.\footnote{We do not report results of the \BARTStages and \PRIMERAStages systems here, since they are not directly trained for the task, and therefore not compatibly comparable. See \autoref{tab:summarizer} and \autoref{tab:recursive-summarizer} in the appendix for full results.}

\autoref{tab:varied_models_results} shows that \ChatGPTSumm outperforms all other methods, including the recursive ChatGPT counterparts. This stresses the advantage of a preliminary selection of aspect-relevant sentences. In addition, it appears that shorter input lengths tend to yield better results, as illustrated by the two size-differences in each of the \PRIMERASumm and \ChatGPTStages models. Finally, the effectiveness of fine-tuning \BARTSumm and \PRIMERASumm with our train set is apparent, as these models are competitive with the relatively strong \ChatGPTSumm model.

We also produced ``Oracle'' extractive summaries for our data, generated greedily to maximize the average of ROUGE-1 and ROUGE-2 scores, given the reference summary \cite{nallapati2017summarunner}.
The large gap between the baseline and Oracle scores, as seen in \autoref{tab:varied_models_results}, leaves much room for improvement in future work on our task.

\begin{table}[t]
\resizebox{\columnwidth}{!}{%
\begin{tabular}{lccc|ccc}
\toprule 
Base Model & \begin{tabular}[c]{@{}c@{}}Input\\ Size\end{tabular} & \begin{tabular}[c]{@{}c@{}}Fine\\ Tuned\end{tabular} & \begin{tabular}[c]{@{}c@{}}Sentence\\ Selector\end{tabular} & R-1           & R-2          & R-L           \\ \midrule 
\BARTSumm                                                                      & 1K                                                   & \cmark                                               & Sent-T5                                                 & 32.4          & 8.3          & 18.7          \\
\PRIMERASumm                                                                   & 4K                                                   & \cmark                                               & Sent-T5                                                 & 30.5          & 8.0          & 18.3          \\
\PRIMERASumm                                                                  & 1K                                                   & \cmark                                               & Sent-T5                                                 & 31.2          & 8.3          & 18.7          \\
\ChatGPTSumm                                                              & 4K                                                   & \xmark                                               & Sent-T5                                                 & \textbf{33.7} & \textbf{9.4} & \textbf{19.8}    \\ 
\ChatGPTStages                                                            & 4K                                                   & \xmark                                               & -                                                       & 32.4          & 9.2          & 19.1          \\
\ChatGPTStages                                                            & 16K                                                  & \xmark                                               & -                                                       & 31.8          & 8.1          & 18.6          \\ \midrule 
Oracle                                                                    &                                                    &                                                     &                                             & 53.4          & 27.8         & 31.7          \\ \bottomrule 
\end{tabular}
}
\caption{The ROUGE $F_1$ scores of different summarization models on our test set. All variants include the aspect label as part of the input.}
\label{tab:varied_models_results}
\end{table}

\begin{table}[t]
\resizebox{\columnwidth}{!}{%
\begin{tabular}{cccc|ccc}
\cmidrule[\heavyrulewidth]{2-7}
 & \begin{tabular}[c]{@{}c@{}}Fine\\ Tuned\end{tabular} & \begin{tabular}[c]{@{}c@{}}Aspect\\ Input\end{tabular}    & \begin{tabular}[c]{@{}c@{}}Sentence\\ Selector\end{tabular} & R-1           & R-2          & R-L           \\ \cmidrule{2-7}
\textcolor{gray}{1}  & \cmark                                               & \cmark & Sent-T5                                                 & \textbf{32.4} & \textbf{8.3} & \textbf{18.7} \\
\textcolor{gray}{2}  & \cmark                                               & \cmark & Lead                                                    & 31.3          & 7.6          & 17.9          \\
\textcolor{gray}{3}  & \cmark                                               & \xmark & Sent-T5                                                 & 32.0          & 8.0          & 18.2          \\
\textcolor{gray}{4}  & \cmark                                               & \xmark & Lead                                                    & 27.5          & 5.3          & 15.7          \\
\textcolor{gray}{5}  & \xmark                                               & \xmark & Lead                                                    & 25.4          & 5.0          & 15.0          \\ \cmidrule{2-7}
\textcolor{gray}{6}  & \cmark                                               & \cmark & \OracleSelector                                                  & 40.6          & 14.5         & 23.2          \\ \cmidrule[\heavyrulewidth]{2-7}
\end{tabular}
}
\caption{ROUGE $F_1$ scores on different configurations of the \BARTSumm model.}
\label{tab:bart_analysis}
\end{table}

\subsection{Ablation Analysis}
\label{subsec:eval_analysis}

We provide insights regarding the contribution of different components in our summarization baselines, operating the \BARTSumm-based models as a use-case (being the best of the two fine-tuned models). We refer to rows in \autoref{tab:bart_analysis} throughout the analysis.

\paragraph{Aspect in input.}
We first observe what happens if the aspect label is left out from the input to the summarizing model.
When using the Sentence-T5 selector, which already selects sentences relevant to the aspect, a very slight improvement in performance is achieved with the aspect in the input (row 1 vs. row 3). However, there is a much larger upgrade when the aspect is input with Lead sentences (row 2 vs. row 4). This indicates that simply providing the requested aspect in the input indeed trains \BARTSumm to attend to aspect-relevant sentences.




\paragraph{Aspect-aware sentence selection.}
Inputting Lead sentences without the aspect is akin to generic multi-document summarization. Row 5 represents this setting without any fine-tuning, i.e., with the original BART model pre-trained for generic summarization. The large difference in scores with respect to the aspect-aware configurations 
suggests that the characteristics of our ABS data is 
distinct from the generic summarization task.



\paragraph{Oracle sentence selection.}
To estimate an upper bound for the \BARTSumm summarizer, we devise an ``Oracle'' that mimics a near optimal sentence selector, denoted \OracleSelector (not to be confused with the Oracle summary in Section \ref{subsec:eval_automatic}). It greedily selects the sentences that maximize ROUGE against the reference aspect-based summary, at the allowed input size limit of $\sim$1K tokens \citep{hayashi-etal-2021-wikiasp} (see Appendix \ref{appendix:Sentence-selector} for details). 
As shown in row 6, using \OracleSelector with \BARTSumm produces substantially greater scores than the next best option, where the Sentence-T5 selector is applied (row 1). This stresses the potential of a good preliminary sentence selector when using the filter-then-summarize approach.

\begin{table}[t]
\resizebox{\linewidth}{!}{%
\begin{tabular}{lcc}
\toprule
\textbf{Model}         & \textbf{\begin{tabular}[c]{@{}c@{}}Relevance to\\  Aspect\end{tabular}} & \textbf{\begin{tabular}[c]{@{}c@{}}Relevance to \\ Reference Summ.\end{tabular}} \\ \midrule
\ChatGPTStages         & \textbf{3.70 (1.38)}                                                     & 2.45 (0.92)                                                                      \\
\ChatGPTSumm - Sent-T5 & 3.40 (1.31)                                                              & \textbf{2.80 (1.01)}                                                              \\
\BARTSumm - Sent-T5           & 3.05 (1.83)                                                             & 2.35 (1.46)                                                                     
                                                  \\ 
\bottomrule
\end{tabular}
}
\caption{Human evaluation results of the three models with the highest ROUGE scores. The evaluation was conducted on 20 system summaries of each model. `Overall' values are the mean (std) scores on a scale of 1--5 for relevance to the aspect and relevance to the respective aspect-based reference summary.}
\label{tab:human_evaluation}
\end{table}

\subsection{Human Evaluation}
\label{subsec:eval_human}

We conduct a manual evaluation on the summaries produced by the three top-scoring models: \ChatGPTStages{}, \ChatGPTSumm{}, and \BARTSumm, with the latter two using the Sentence-T5 sentence selector.
For 20 random instances from the \openasp{} test set, we assessed: (1) Relevance to the aspect, i.e., ``is the target aspect adequately discussed in the system summary?''; and (2) Relevance to the Reference summary, i.e., ``does the system summary refer to the information in the aspect-based reference summary?'' \citep{ernst-etal-2022-proposition, lebanoff-etal-2018-adapting}. Each criterion was rated on a 1--5 scale, 5 being best. The outcomes of the evaluation are presented in \autoref{tab:human_evaluation}. 

The ranking of relevance to the reference summary is consistent with the automatic score ranking (in \autoref{tab:varied_models_results}). Importantly, the manual scores achieved are quite low (all 2--3). Furthermore, the models demonstrate varying levels of success in extracting aspect-relevant information,
as indicated by the moderate aspect-relevancy scores and high standard deviations. Overall, these observations re-emphasize the
challenges posed for models on the task.

\section{Conclusion}

Summarizing texts around an open-aspect is a basic necessity when consuming information. 
Our new \openasp{} benchmark serves this demand, as the first \textit{open} ABS dataset in the \textit{multi}-document setting, with \textit{high}-quality summaries collected via an efficient protocol. Our protocol overcomes the major hurdle of manually collecting summaries, by tapping into existing generic summaries in multi-document summarization datasets. Our proposed baselines, based on strong models, reveal the gap towards solving this task, posing a challenge even for the best current models. Overall, our efficient data collection protocol can be expanded to supply even more data for real-world open-ABS and related information-seeking tasks.




\section*{Limitations}
\label{sec_limitations}
This study leverages existing generic multi-document summaries to generate aspect-based summaries by manually extracting aspect-related sentences. While this approach proved effective for the specific news datasets we used, it may not be readily applicable to different datasets where aspect-related sentences from the summary may not accurately capture all the necessary information for that aspect. We assess this in our analyses, and recommend to do so on other potential datasets on which our protocol is applied.

Although \openasp{} contains a representative sample of aspects from the generic summaries, the overall distribution of aspect labels is sparse with a small fraction of repeating labels. This limits further analysis of aspect distribution or aspect discovery that we leave for future work.

Furthermore, the usage of ChatGPT raises certain concerns despite its popularity. Firstly, the lack of detailed documentation regarding ChatGPT's training procedure makes it challenging to determine the specific training-data used. This raises the possibility of contamination, where our test data might have been incorporated somehow into the training of ChatGPT.

Finally, for our experiments, we employed specific prompts (detailed in the Appendix) to assess the capabilities of ChatGPT for our task. Although we attempted several prompts, it is important to note that other prompts could yield different outputs. Consequently, we cannot make definitive claims about the model's capabilities.

\section{Ethics and Broader Impact} 

This paper is submitted in the wake of a tragic terrorist attack perpetrated by Hamas, which has left our nation profoundly devastated. On October 7, 2023, thousands of Palestinian terrorists infiltrated the Israeli border, launching a brutal assault on 22 Israeli villages. They methodically moved from home to home brutally torturing and murdering more than a thousand innocent lives, spanning from infants to the elderly. In addition to this horrifying loss of life, hundreds of civilians were abducted and taken to Gaza. The families of these abductees have been left in agonizing uncertainty, as no information, not even the status of their loved ones, has been disclosed by Hamas.

The heinous acts committed during this attack, which include acts such as shootings, sexual assaults, burnings, and beheadings, are beyond any justification.

In addition to the loss we suffered as a nation and as human beings due to this violence, many of us feel abandoned and betrayed by members of our research community who did not reach out and were even reluctant to publicly acknowledge the inhumanity and total immorality of these acts.

We fervently call for the immediate release of all those who have been taken hostage and urge the academic community to unite in condemnation of these unspeakable atrocities committed by Hamas, who claim to be acting in the name of the Palestinian people. We call all to join us in advocating for the prompt and safe return of the abductees, as we stand together in the pursuit of justice and peace.

\section{Acknowledgements}\label{sec:acknowledgements}
This work was supported by the Israel Science Foundation (grant no. 2827/21), the Israel Ministry of Science and Technology, and One AI.



\bibliography{anthology,custom}
\bibliographystyle{acl_natbib}


\appendix

\begin{figure*}[ht]
    \centering
    \includegraphics[width=\linewidth]{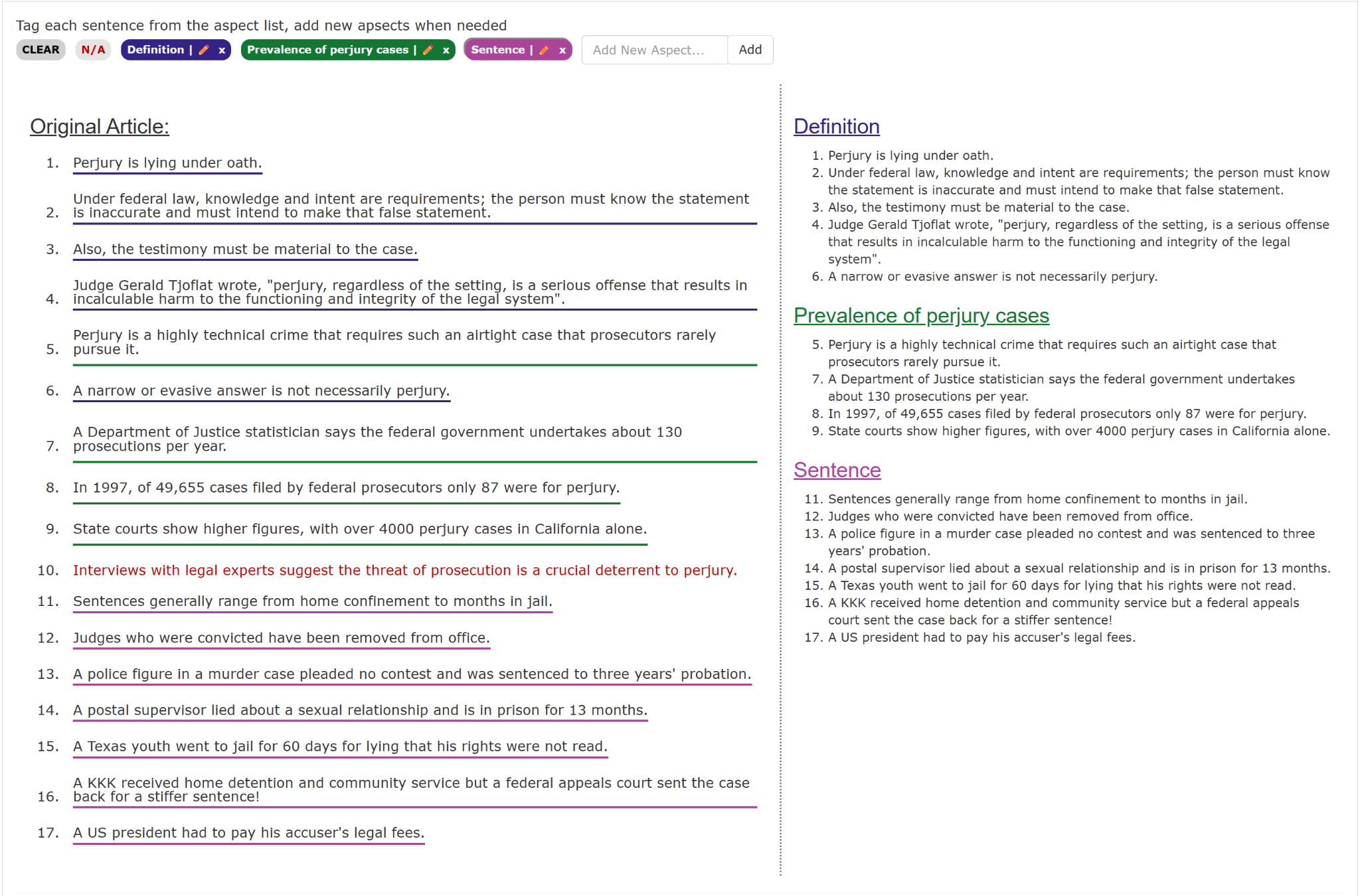}
    \caption{Our interface for annotating aspect-based summaries. An annotator can add, remove or edit aspect labels (top), and respectively select sentences in the source reference summary that are relevant to the aspects. The produced summaries shown on the right side and updated in real time. Sentences with no aspect labels should actively annotated with special "N/A" label (sentence 10) to ensure workers reading all the article content.}
    \label{fig:annotation_ui_full}
\end{figure*}
\begin{figure*}[t]
    \centering
    \includegraphics[scale=0.4]{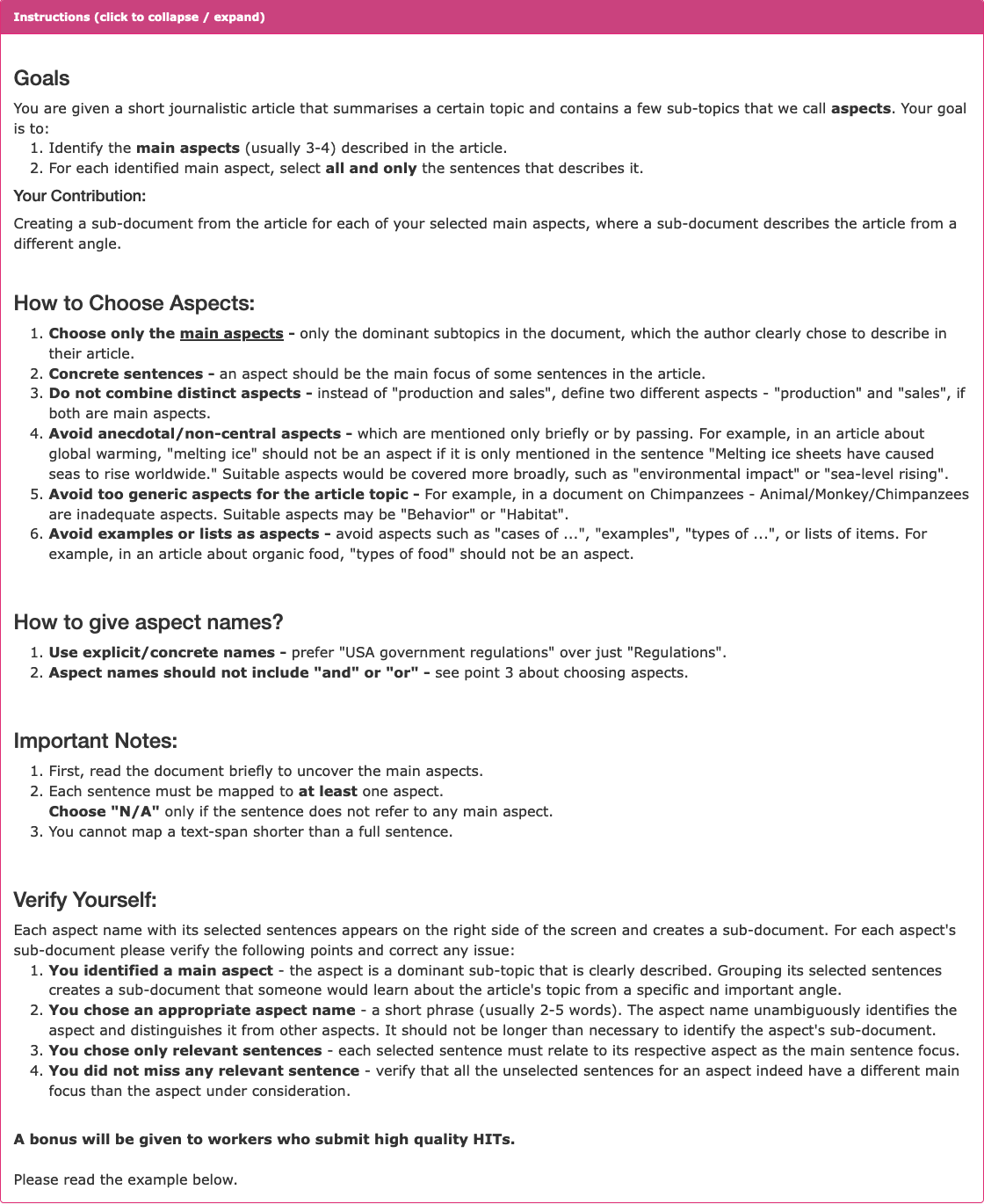}
    \caption{Our annotation guidelines for extracting aspects and respective aspect-based summaries, as shown to crowdworkers on the Mechanical Turk platform.}
    \label{fig:ui_guidelines}
\end{figure*}
\begin{figure*}[t]
    \centering
    \includegraphics[scale=0.4]{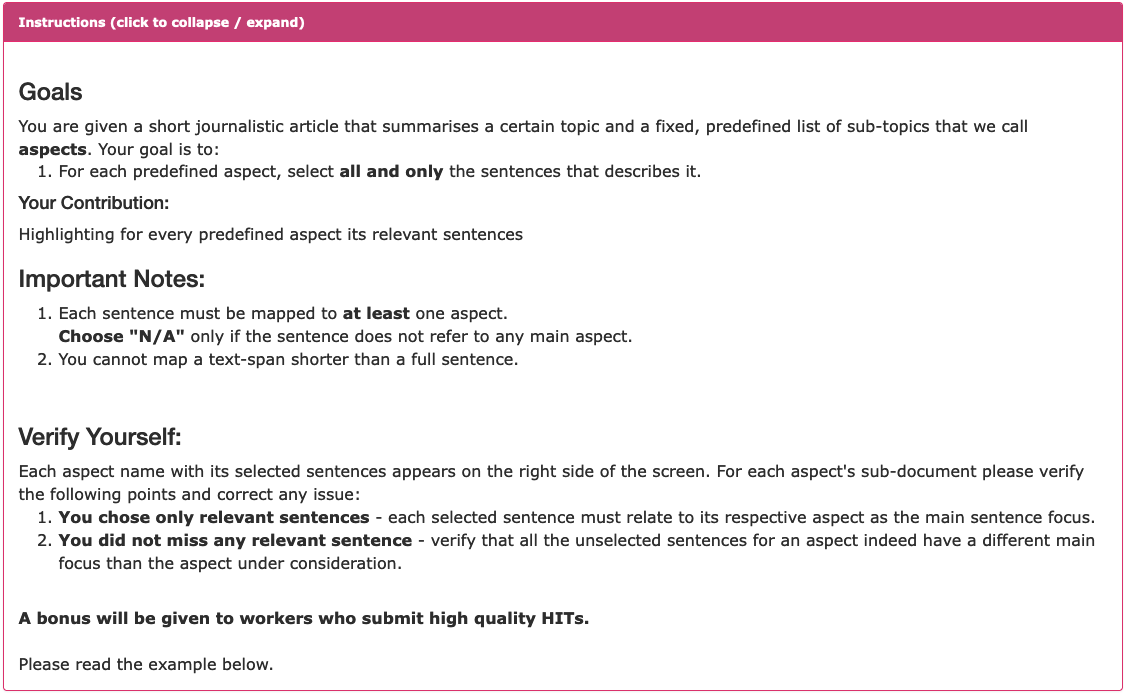}
    \caption{Our annotation guidelines for aligning document-sentences to an aspect (Section \ref{subsec:dataset-eval}), as presented to crowdworkers on the Mechanical Turk platform.}
    \label{fig:ui_src_guidelines}
\end{figure*}

\section{\openasp{} Collection Details}\label{appendix:annotation-guidelines}

\subsection{Annotation Interfaces}
We present screenshots of the UI and instructions to extract aspect-based summaries for our dataset in Figures \ref{fig:annotation_ui_full} and \ref{fig:ui_guidelines}, respectively. The provided instructions assist annotators in accurately identifying aspects and extracting relevant information related to each of them. 

Furthermore, we show in Figure \ref{fig:ui_src_guidelines} a screenshot with the instructions provided to extract aspect-related source sentences for the analysis-valid and analysis-test sets. 
The UI was similar to the version we used for creating our OpenAsp dataset except for the given aspects that couldn't be edited or selected as in OpenAsp. The annotators only selected the related sentences to each predefined aspect from a given source document.  

\begin{figure*}[t]
    \centering
    \includegraphics[scale=0.4]{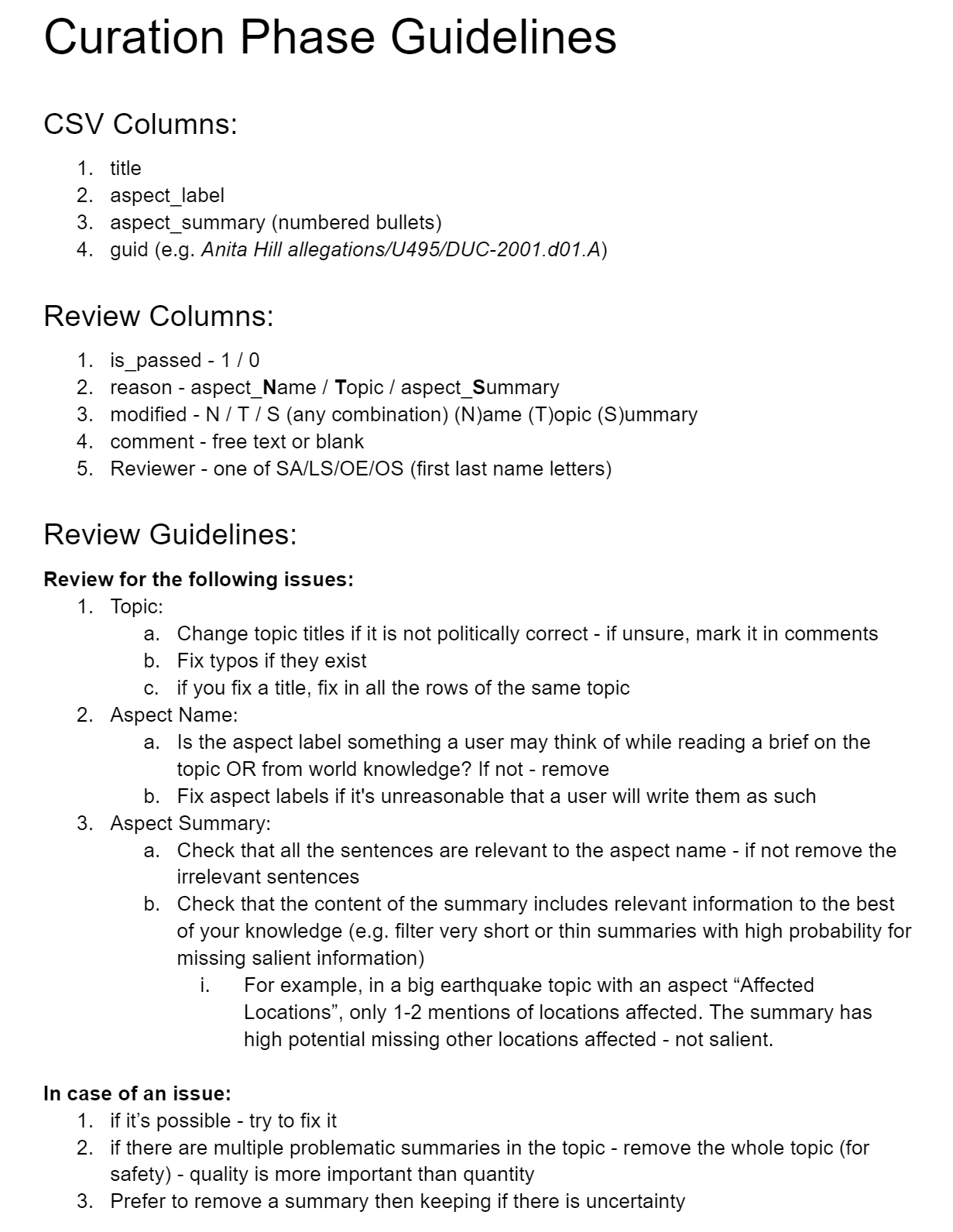}
    \caption{Curation phase guidelines used for cleaning \openasp{}.}
    \label{fig:curation_guidelines}
\end{figure*}
\begin{table*}[ht]
\centering
\small
\resizebox{\textwidth}{!}{%
\begin{tabular}{c|c|l|c|c|c|c|l}
\hline
\textbf{Topic}                                                   & \textbf{\begin{tabular}[c]{@{}c@{}}Original \\ aspect label\end{tabular}}     & \multicolumn{1}{c|}{\textbf{Original summary}}                                                                                                                                                                                                                                                                                                                                                                                                                                                                                                                                                 & \textbf{\begin{tabular}[c]{@{}c@{}}Edited \\ aspect label\end{tabular}} & \textbf{\begin{tabular}[c]{@{}c@{}}Is \\ passed\end{tabular}} & \textbf{Reason} & \textbf{Modified}                                     & \multicolumn{1}{c}{\textbf{Comment}}                                                                                                 \\ \hline
\begin{tabular}[c]{@{}c@{}}Earthquakes \\ magnitude\end{tabular} & \begin{tabular}[c]{@{}c@{}}Destructive \\ earthquakes \\ below 7\end{tabular} & \begin{tabular}[c]{@{}l@{}}9.   Two quakes caused great destruction even \\       though their magnitudes were below the 7 level. \\ 10. Twenty-five thousand were killed in Armenia, \\       and at least another 1,000 in Tadzhikistan.\end{tabular}                                                                                                                                                                                                                                                                                                                                        & -                                                                       & \xmark                                                             & Topic           & -                                                     & \begin{tabular}[c]{@{}l@{}}The topic is a \\ list of incidents \\ and therefore \\ all its \\ summaries \\ are rejected\end{tabular} \\ \hline
\begin{tabular}[c]{@{}c@{}}Clarence \\ Thomas\end{tabular}       & \begin{tabular}[c]{@{}c@{}}Law \\ career\end{tabular}                         & \begin{tabular}[c]{@{}l@{}}0.   Clarence Thomas, a black conservative republican, \\       was confirmed as a Supreme Court justice on \\       16 October 1991. \\ 8.   He worked as an attorney and moved to Washington \\       in 1974. \\ 9.   Thomas was appointed to several civil rights and \\       equal employment opportunity positions beginning \\       in the early 1980s and as a judge in the U.S. \\ 10. Circuit Court of Appeals in 1990. \\ \textbf{12. His confirmation as a justice of the Supreme Court} \\       \textbf{brought joy to his mother in Pinpoint, Georgia.}\end{tabular} & -                                                                       & \cmark                                                             & -               & Summary                                               & \begin{tabular}[c]{@{}l@{}}Sentence 12 \\ is not relevant \\ and should \\ be removed.\end{tabular}                                  \\ \hline
Autism                                                           & \begin{tabular}[c]{@{}c@{}}People \\ with \\ autism\end{tabular}              & \begin{tabular}[c]{@{}l@{}}19. Autism affects 1 in 500-1000 and is on the rise. \\ 20. Eighty percent are boys. \\ 21. Odds are 1 in 20 that a family with one autistic \\       child will have another. \\ 22. Brick, NJ has an autism cluster.\end{tabular}                                                                                                                                                                                                                                                                                                                                 & \begin{tabular}[c]{@{}c@{}}Autism \\ prevalence\end{tabular}            & \cmark                                                             & -               & \begin{tabular}[c]{@{}c@{}}Aspect\\ name\end{tabular} & \begin{tabular}[c]{@{}l@{}}The current \\ name is \\ ambiguous \\ and therefore \\ is changed\end{tabular}                           \\ \hline
\end{tabular}}
\caption{Examples of three data instances that were modified or rejected during the curation phase. \textbf{Bold} text represents a problematic sentence in the summary. The `Is passed' column states whether the instance is included in the final data or is rejected. `Reason' explains why the instance was rejected. `Modified' specifies the type of problem (`Aspect name', `Summary', or `Topic') in case the instance needs a correction in order to be included in the final dataset.}
\label{tab:curation_examples}
\end{table*}

\subsection{Curation Phase Details}\label{app:dataset-curation-details}
Four paper authors applied the curation phase as detailed in Section \ref{sec:curation-phase}. We first defined and refined the curation guidelines (\autoref{fig:curation_guidelines}) in a discussion before starting curation.
Then, we split the data evenly among curators, by original MTurk annotators and by data source splits (DUC vs. Multi-News) to avoid personal biases in specific sub-parts of the data. Some curated samples are shown in \autoref{tab:curation_examples}. The released \openasp{} dataset includes the original aspect labels and summaries, as well as the instances from after curation.

\section{Models Implementation Details}\label{appendix:implementation_details}

\begin{figure*}[t]
\centering
\resizebox{0.9\textwidth}{!}{%
\includegraphics[]{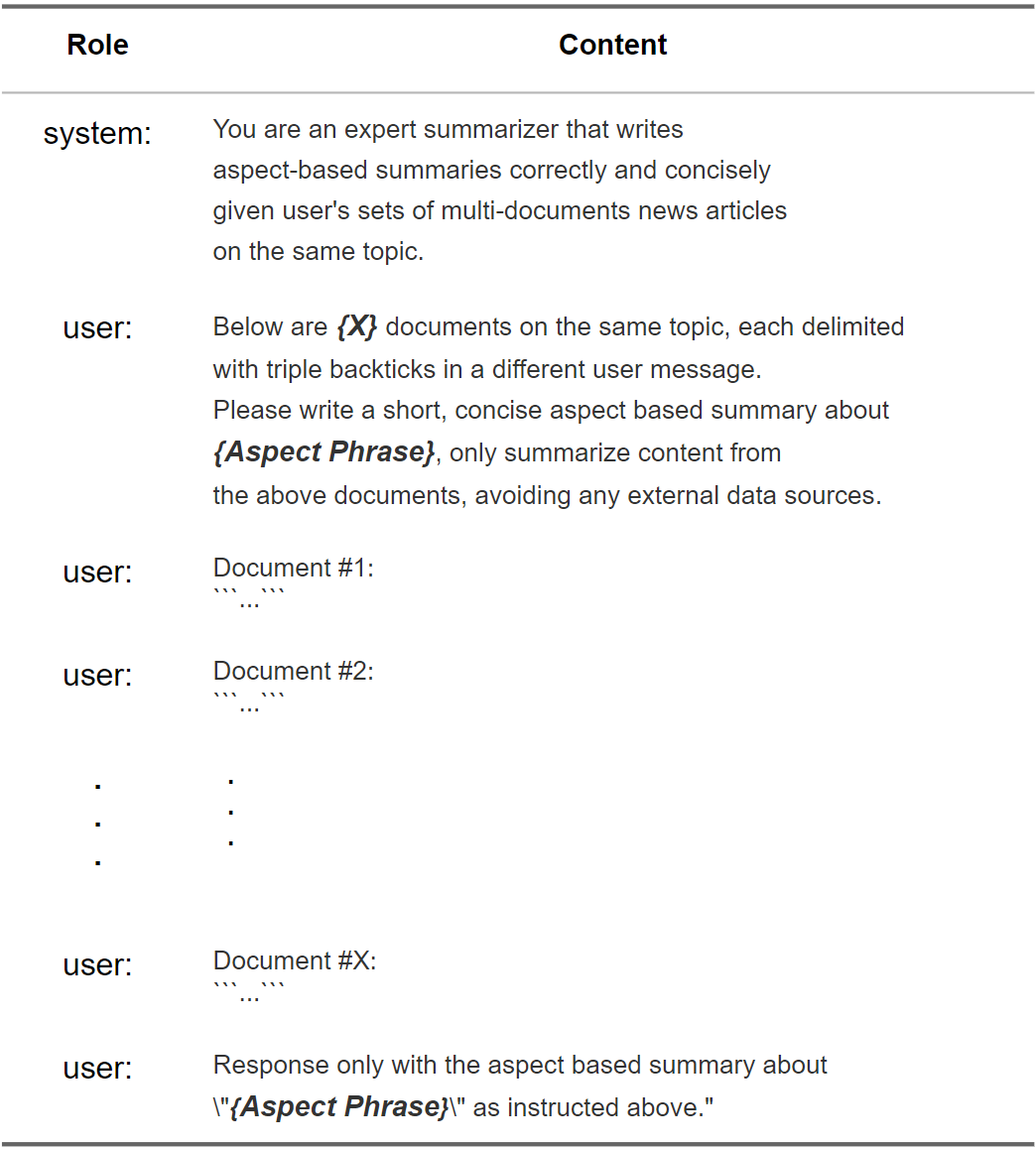}
}
    \caption{The prompt template used for all ChatGPT models, \ChatGPTSumm, \ChatGPTStages and \ChatGPTStageslong. In the recursive configuration, when summarizing summaries, the summaries act as ``documents''.}
    \label{appendix:chat-gpt_prompt_fig}
    
\end{figure*}

\subsection{Fine-tuned Models}
\paragraph{BART.} We used a variant of the BART-large model with 406M parameters, fine-tuned on the CNN Daily Mail dataset\footnote{https://huggingface.co/facebook/bart-large-cnn}.

\paragraph{PRIMERA.} We utilized a variant of PRIMERA which was fine-tuned on the MultiNews dataset\footnote{https://huggingface.co/allenai/PRIMERA-multinews} and includes 447M parameters

We fine-tuned PRIMERA and BART on \openasp{} train set on 2 V100 GPUs with the following hyper-parameters: learning rate of 10e-5, batch size of 1, gradient accumulation steps of 3, and 3 epochs. 

\subsection{ChatGPT}\label{appendix:chatgpt-details}
In Section \ref{appendix:chatgpt-details}, we experimented with three models based on OpenAI's \emph{ChatGPT} API - \ChatGPTSumm, \ChatGPTStages based on \texttt{gpt-3.5-turbo-0301}, while \ChatGPTStageslong uses \texttt{gpt-3.5-turbo-16k-0613}.
The temperature of all models is set to 0 for reproducibility.


To determine the appropriate prompt, we manually evaluated the aspect-based summaries generated during a brief manual tuning of the prompt text. The final prompt used for all ChatGPT models is presented in Figure \ref{appendix:chat-gpt_prompt_fig}.

In this paper, we used ChatGPT for summarization in two approaches, \ChatGPTSumm and \ChatGPTStages.
\ChatGPTSumm summarizes a reduced version of the original documents that fits within the model's input length limit. The reduced version is created using a sentence selection method (described in Section \ref{subsec:decomposed-models}).
The selected sentences from each document in the document set are then consecutively presented within the same prompt as individual entries
(\emph{document\#1: ```...```, ... , document\#X: ```...```}).

For \ChatGPTStages and \ChatGPTStageslong models,
full documents are concatenated (separated by document title \emph{document\#1:}) until reaching the input length limit. Then the model generates a summary for the first portion of sentences. This process is repeated until all documents are summarized once. Then, the model gets all summaries with the same prompt, and summarizes the summaries to produce the final summary.



\subsection{Oracle Sentence-Selector Details}
\label{appendix:Sentence-selector}
Following \citet{hayashi-etal-2021-wikiasp} the oracle sentence selector is implemented as follows. Sentences from the document sets are aggregated to maximize the average of ROUGE-1 and ROUGE-2 F-1 scores against the reference aspect-based summaries. Once the score no longer improves, the selected sentences are omitted from the source, and the process is repeated again. This continues until the input size limit is reached, e.g., 1K tokens for BART.

\section{Model results}

\begin{table*}[ht]
\centering
\resizebox{\textwidth}{!}{%
\begin{tabular}{cccl|rrr|rrr|rrr}
\toprule
\multirow{2}{*}{\begin{tabular}[c]{@{}c@{}}Fine\\ Tuned\end{tabular}} & \multirow{2}{*}{\begin{tabular}[c]{@{}c@{}}Aspect\\ Input\end{tabular}} & \multirow{2}{*}{\begin{tabular}[c]{@{}c@{}}Input\\ Size\end{tabular}} & \multicolumn{1}{l|}{\multirow{2}{*}{\begin{tabular}[c]{@{}c@{}}Base Model\end{tabular}}} & \multicolumn{3}{c|}{{\ul OOracle Selector}}                                                        & \multicolumn{3}{c|}{{\ul LLead Selector}}                                                         & \multicolumn{3}{c}{{\ul SS-T5 Selector}}                                                         \\
                                                                     &                                                                       &                                                                       & \multicolumn{1}{c|}{}                                                                       & \multicolumn{1}{l}{R-1}  & R-2                               & R-L                                & \multicolumn{1}{l}{R-1}  & R-2                              & R-L                                & \multicolumn{1}{l}{R-1}  & R-2                              & R-L                               \\ \midrule
\cmark                                                               & \cmark                                                                    & 1K                                                                & \BARTSumm                                                                                        & \textbf{40.6}            & \multicolumn{1}{c}{\textbf{14.5}} & \multicolumn{1}{c|}{\textbf{23.2}} & 31.3                     & \multicolumn{1}{c}{7.6}          & \multicolumn{1}{c|}{17.9}          & 32.4                     & \multicolumn{1}{c}{8.3}          & \multicolumn{1}{c}{18.7}          \\
\cmark                                                               & \cmark                                                                    & 4K                                                                & \PRIMERASumm                                                                                     & \multicolumn{1}{l}{33.8} & 10.0                              & 20.0                               & \multicolumn{1}{l}{30.0} & 7.6                              & 17.9                               & \multicolumn{1}{l}{30.5} & 8.0                              & 18.3                              \\
\cmark                                                               & \cmark                                                                    & 1K                                                                & \PRIMERASumm                                                                                     & \multicolumn{1}{l}{37.9} & 13.0                              & 22.6                               & \multicolumn{1}{l}{29.9} & 7.6                              & 17.9                               & \multicolumn{1}{l}{31.2} & 8.3                              & 18.7                              \\ \hline
\cmark                                                               & \xmark                                                                    & 1K                                                                & \BARTSumm                                                                                        & 39.5                     & \multicolumn{1}{c}{13.3}          & \multicolumn{1}{c|}{22.5}          & 27.5                     & \multicolumn{1}{c}{5.3}          & \multicolumn{1}{c|}{15.7}          & 32.0                     & \multicolumn{1}{c}{8.0}          & \multicolumn{1}{c}{18.2}          \\
\cmark                                                               & \xmark                                                                    & 4K                                                                & \PRIMERASumm                                                                                     & \multicolumn{1}{l}{31.4} & 8.2                               & 18.6                               & \multicolumn{1}{l}{26.4} & 5.2                              & 15.7                               & \multicolumn{1}{l}{29.1} & 6.6                              & 17.0                              \\
\cmark                                                               & \xmark                                                                    & 1K                                                                & \PRIMERASumm                                                                                     & \multicolumn{1}{l}{36.6} & 12.4                              & 21.9                               & \multicolumn{1}{l}{26.6} & 5.2                              & 15.6                               & \multicolumn{1}{l}{30.4} & 7.5                              & 17.8                              \\ \hline
\xmark                                                               & \cmark                                                                    & 1K                                                                & \BARTSumm                                                                                        & 34.6                     & \multicolumn{1}{c}{10.7}          & \multicolumn{1}{c|}{20.5}          & 25.5                     & \multicolumn{1}{c}{5.0}          & \multicolumn{1}{c|}{15.3}          & 28.5                     & \multicolumn{1}{c}{6.7}          & \multicolumn{1}{c}{17.0}          \\
\xmark                                                               & \cmark                                                                    & 4K                                                                & \PRIMERASumm                                                                                     & 30.9                     & \multicolumn{1}{c}{7.5}           & \multicolumn{1}{c|}{17.3}          & 28.7                     & \multicolumn{1}{c}{6.3}          & \multicolumn{1}{c|}{16.0}          & 29.5                     & \multicolumn{1}{c}{6.6}          & \multicolumn{1}{c}{16.4}          \\
\xmark                                                               & \cmark                                                                    & 1K                                                                & \PRIMERASumm                                                                                     & \multicolumn{1}{l}{35.8} & 11.1                              & 20.0                               & \multicolumn{1}{l}{29.0} & 6.4                              & 16.2                               & \multicolumn{1}{l}{31.3} & 7.8                              & 17.6                              \\
\xmark                                                               & \cmark                                                                    & 4K                                                                & \ChatGPTSumm                                                                                & 35.6                     & \multicolumn{1}{c}{10.8}          & \multicolumn{1}{c|}{20.7}          & \textbf{32.8}            & \multicolumn{1}{c}{\textbf{8.9}} & \multicolumn{1}{c|}{\textbf{19.2}} & \textbf{33.7}            & \multicolumn{1}{c}{\textbf{9.4}} & \multicolumn{1}{c}{\textbf{19.8}} \\ \hline
\xmark                                                               & \xmark                                                                    & 1K                                                                & \BARTSumm                                                                                        & 34.4                     & \multicolumn{1}{c}{10.6}          & \multicolumn{1}{c|}{20.2}          & 25.4                     & \multicolumn{1}{c}{5.0}          & \multicolumn{1}{c|}{15.0}          & 28.2                     & \multicolumn{1}{c}{6.5}          & \multicolumn{1}{c}{16.9}          \\
\xmark                                                               & \xmark                                                                    & 4K                                                                & \PRIMERASumm                                                                                     & 31.1                     & \multicolumn{1}{c}{7.7}           & \multicolumn{1}{c|}{17.4}          & 28.5                     & \multicolumn{1}{c}{6.2}          & \multicolumn{1}{c|}{16.1}          & 29.6                     & \multicolumn{1}{c}{6.7}          & \multicolumn{1}{c}{16.7}          \\
\xmark                                                               & \xmark                                                                    & 1K                                                                & \PRIMERASumm                                                                                     & \multicolumn{1}{l}{35.8} & 11.0                              & 20.0                               & \multicolumn{1}{l}{28.9} & 6.3                              & 16.0                               & \multicolumn{1}{l}{31.1} & 7.5                              & 17.4                              \\ \bottomrule
\end{tabular}
}
\caption{The ROUGE $F_1$ scores for all model configurations that use a sentence selector. The number of extracted sentences by a sentence selector is limited by the maximum input token-length, as indicated in `Input Size'. The best configuration for each sentence selector option is marked in \textbf{bold}.}
\label{tab:summarizer}
\end{table*}

\begin{table}[t]
\resizebox{\columnwidth}{!}{%
\begin{tabular}{lccc|ccc}
\toprule 
Base Model & \begin{tabular}[c]{@{}c@{}}Input\\ Size\end{tabular} & \begin{tabular}[c]{@{}c@{}}Fine\\ Tuned\end{tabular} & \begin{tabular}[c]{@{}c@{}}Sentence\\ Selector\end{tabular} & R-1           & R-2          & R-L           \\ \midrule
\BARTStages                                                                      & 1K                                                   & \cmark                                               & \OracleSelector                                                 & 29.8          & 6.3          & 16.8          \\
\PRIMERAStages                                                                   & 4K                                                   & \cmark                                               & \OracleSelector                                                 & 28.6          & 6.4          & 16.9          \\
\PRIMERAStages                                                                   & 1K                                                   & \cmark                                               & \OracleSelector                                                 & 27.3          & 5.6          & 16.2 \\ \midrule

\BARTStages                                                                      & 1K                                                   & \cmark                                               & Sent-T5                                                 & 29.1          & 6.0          & 16.2          \\
\PRIMERAStages                                                                   & 4K                                                   & \cmark                                               & Sent-T5                                                 & 29.1          & 6.7          & 17.1          \\
\PRIMERAStages                                                                   & 1K                                                   & \cmark                                               & Sent-T5                                                 & 27.9          & 5.7          & 16.4
\\ \midrule
\BARTStages                                                                      & 1K                                                   & \cmark                                               & Lead                                                 & 30.1          & 6.6          & 17.1          \\
\PRIMERAStages                                                                   & 4K                                                   & \cmark                                               & Lead                                                 & 29.8          & 7.3          & 17.8          \\
\PRIMERAStages                                                                   & 1K                                                   & \cmark                                               & Lead                                                 & 28.1          & 5.8          & 16.4 \\ \midrule
\ChatGPTStages                                                            & 4K                                                   & \xmark                                               & -                                                       & \textbf{32.4}          & \textbf{9.2}          & \textbf{19.1}          \\
\ChatGPTStages                                                            & 16K                                                  & \xmark                                               & -                                                       & 31.8          & 8.1          & 18.6          \\ \bottomrule 
\end{tabular}
}
\caption{The ROUGE $F_1$ scores for all model configurations in the Recursive summarization technique. All variants include the aspect label as part of the input. \BARTStages and \PRIMERAStages were fine-tuned using the Filter-then-Summarize approach (on the input sentences extracted by a sentence selector), however executed through the recursive technique (see Section \ref{subsec:end2end-models}). The best configuration is marked in \textbf{bold}.}
\label{tab:recursive-summarizer}
\end{table}
\autoref{tab:summarizer}  presents the complete set of experiments conducted on our baseline models in the Filter-then-Summarize technique, employing the Lead, Sentence-T5, and Oracle sentence selectors.

\autoref{tab:recursive-summarizer} presents the experiments conducted on our baseline models in the Recursive summarization technique. As a reminder from Section \ref{subsec:end2end-models}, \BARTStages and \PRIMERAStages were fine-tuned with the Filter-then-Summarize approach, which can use three different sentence selectors (Lead, Sent-T5 and \OracleSelector). In \autoref{tab:recursive-summarizer} it is apparent that the `Lead' selector provides superior results over the other two sentence selectors. `Lead' sentences are less focused on aspect-specific information (as revealed in Section \ref{subsec:eval_automatic}). We can hence assume that this characteristic encourages the model to focus more on the aspect during training, and consequently to perform better during inference.

\subsection{System Summary Examples}\label{appendix:models_summaries_examples}
Tables \ref{tab:orbit} and \ref{tab:reason_unemployment} present the aspect-based summaries generated by varied models for 2 different aspects, \emph{'Launch into orbit'} and \emph{'Reasons for high unemployment rates'}, respectively. The corresponding reference summaries appear in the button line.

\section{\openasp\ Details}
\label{sec:appendix_dataset_details}

\subsection{\openasp{} Source Splits}\label{app:dataset-source-splits}

\begin{table}[t]
\resizebox{\columnwidth}{!}{%
\begin{tabular}{c|lrr}
\toprule
Split                 & Source      & \# Topics & \# Aspects \\ \midrule
\multirow{4}{*}{Test} & DUC-2002    & 56        & 157          \\
                      & DUC-2007    & 42        & 148          \\
                      & MultiNews-Test  & 94        & 291          \\
                      & \textbf{Total}       & \textbf{192}       & \textbf{596}          \\ \midrule
\multirow{4}{*}{Valid} & DUC-2001-Test    & 26        & 78           \\
                      & DUC-2006    & 13        & 38           \\
                      & MultiNews-Valid & 43        & 122          \\
                      & \textbf{Total}       & \textbf{82}        & \textbf{238}          \\ \midrule
\multirow{4}{*}{Train} & DUC-2001-Train    & 28        & 82           \\
                      & DUC-2006    & 34        & 115          \\
                      & MultiNews-Train & 83        & 279          \\
                      & \textbf{Total}       & \textbf{145}       & \textbf{476}         \\
\bottomrule
\end{tabular}}
\caption{The size of the \openasp{} dataset splits, broken down to their sources. \# Topics denotes the number of document sets in the split; \# Aspects is the total number of aspect-based summaries in the split.}
\label{tab:dataset-splits-count-full}
\end{table}

\begin{figure*}[t]
    \centering
    \includegraphics[width=0.7\linewidth]{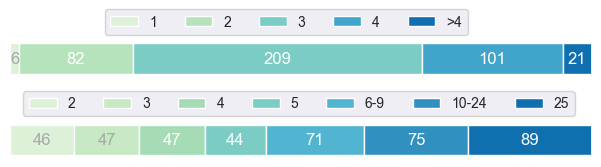}
    \caption{The distribution of the number of aspects (top) and number of documents (bottom) per topic for \openasp{}'s 419 topics. \openasp{} has an average of 3.1 aspects and 10.4 documents per topic.}
    \label{fig:aspect-counts-pie}
\end{figure*}

\begin{figure*}[t]
    \centering
    \includegraphics[width=0.7\linewidth]{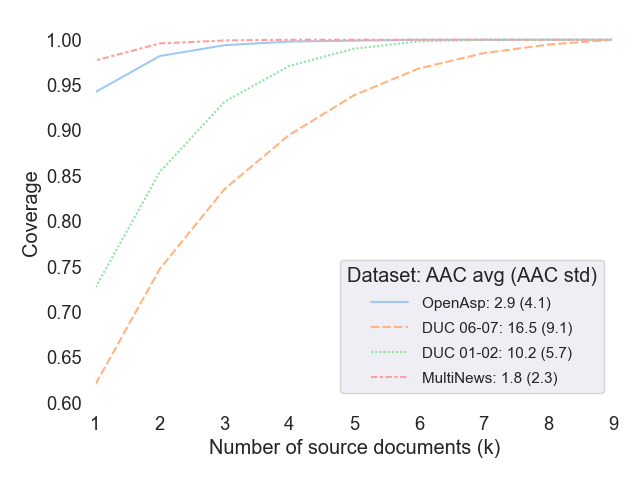}
    \caption{The multi-document coverage and the \textit{dispersion score} (AAC) of \openasp{} compared to other MDS datasets as reported in \citet{wolhandler-etal-2022-multi}. A larger area above the curve implies that summaries rely on a larger subset of their corresponding document-set.}
    \label{fig:multimds-plot}
\end{figure*}

We split \openasp{} dataset into train, validation and test sets based on the source datasets they originated from (see \autoref{tab:dataset-splits-count-full}).
For aspect-based summaries originated from MultiNews, we followed the original splits, for summaries originated from DUC, we separated the test years from the train and validation.

\subsection{MultiNews Filtering}\label{appendix:data-sources}

During annotation phase, we noticed several faulty source documents from MultiNews, probably due to failed crawling. We manually examine some suspicious short source documents, finding a few common phrases that imply the document retrieval failed. For example, \textit{``The seed for this crawl was a list of every host in the Wayback Machine
This crawl was run at a level 1 (URLs including their embeds, plus the URLs of all outbound links including their embeds)
The WARC files associated with this crawl are not currently available to the general public."} and several similar cases.

We created a short list of such texts and automatically filtered all topics from MultiNews containing one or more source documents matching these strings. 

\subsection{Documents and Summaries Lengths}
\begin{figure*}[ht]
    \centering
    \includegraphics[width=\linewidth]{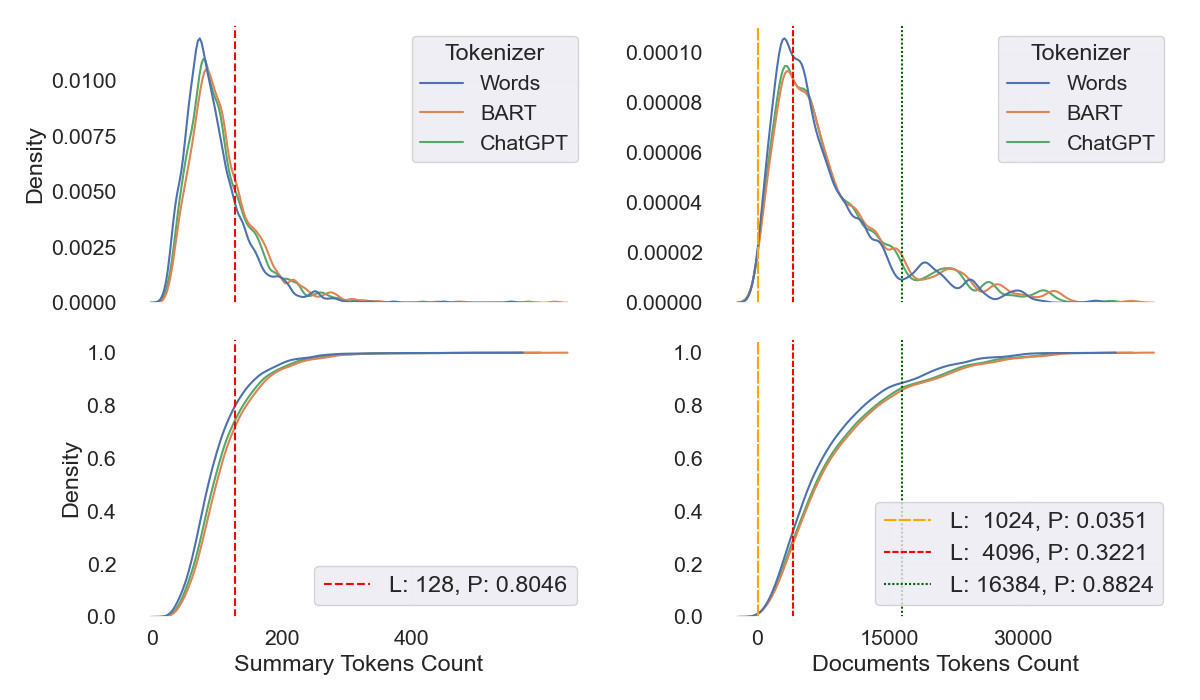}
    \caption{
    The distribution (top) and cumulative distribution (bottom) of token-lengths in summaries (left) and document-sets (right) in \openasp{}. Each plot is presented with three different tokenization methods: standard word tokenization (NLTK word\_tokenize), BART (also used by PRIMERA) and the ChatGPT tokenizer (\href{https://github.com/openai/tiktoken}{tiktoken}). The dashed horizontal lines mark 1024, 4096 and 16384 token lengths. For each horizontal line, $P$ is the percentage of instances in the dataset whose summary/document-set lengths are less than or equal to the corresponding length limit ($L$), using the word tokenizer. For example, 32.21\% of the instances have input word-lengths less than 4,096 words. As seen, the differences between the tokenizers in terms of content length are negligible.}
    \label{fig:data-lengths-full}
\end{figure*}

Figure \ref{fig:data-lengths-full}, presents the distribution and cumulative distribution of token lengths in summaries and
document-sets.

\subsection{Content Diversity}
\label{sec:appendix_content_diversity}
\begin{figure*}[ht]
    \centering
    \includegraphics[width=0.5\linewidth]{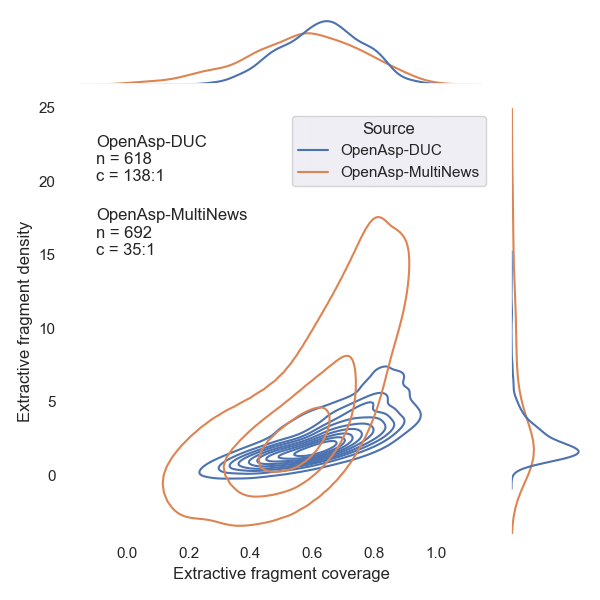}
    \caption{The content diversity graphs \cite{grusky-etal-2018-newsroom} of \openasp{}, separated by the source datasets DUC and MultiNews (a stratified version of \autoref{fig:content_diversity_main}). $n$ and $c$ denote number of examples and median compression ratio respectively. DUC-based summaries contains extracts that are substantially shorter than those of MultiNews.}
    \label{fig:content_diversity_by_source}
\end{figure*}

Content diversity \cite{grusky-etal-2018-newsroom, fabbri-etal-2019-multi} is a joint measure for extractiveness of \emph{coverage} and \emph{density}. \emph{Coverage} is the percentage of words in the summary that are from the source article, and \emph{density} is the average length of the extractive fragment to which each summary word belongs.
While \openasp{}'s summaries are extracted from source generic summaries, the generic summaries themselves are at different levels of abstractiveness relative to their corresponding document set. Accordingly, the aspect-based summaries also exhibit varying abstractive extents. \autoref{fig:content_diversity_main} illustrates the distribution of coverage on the x-axis, and that of density on the y-axis. As shown, most extracts are short, however there are few cases of sentence-level extractions. Overall, there is a diversified balance of abstractiveness in the data.
Figure \ref{fig:content_diversity_by_source}, presents the content diversity graphs of \openasp, separated by the source datasets DUC and MultiNews (a stratified version of Figure \ref{fig:content_diversity_main}).

\subsection{Multi-document Coverage}
Multi-document coverage quantifies the amount of documents in the document-set upon which a corresponding summary depends \citep{wolhandler-etal-2022-multi}.
\autoref{fig:multimds-plot} shows the multi-document coverage of \openasp{}, compared to the source datasets. The coverage is derived in terms of propositional alignment \citep{ernst-etal-2022-proposition} between the highest-matching subset of $k$ documents (x-axis) and the summary.
The \textit{dispersion score}, a function of the area-above-the-curve, is a measure of multi-document coverage, where a higher value means there is higher document-diversity overall.

The \textit{dispersion score} of \openasp{} is 2.9, with standard deviation of 4.1, while DUC-2001/2 and DUC-2006/7 render higher dispersion scores of 10.2 and 16.5 respectively. This can be explained by the longer summaries in the latter datasets, $\sim$2.5 times longer than \openasp{} summaries.
Moreover, a generic summary, as opposed to an aspect-based summary, is expected to cover several subtopics, and hence align with a wider range of source documents.
Meanwhile, the high standard deviation in the dispersion score means there are still aspect-based summaries that align with a larger number of documents. Some summaries require handpicking information from specific documents, while others require consolidating information from across the document set.

\subsection{Adding Topic Names}
We add to each document-set in our dataset a topic name, which was selected as an additional side task. Specifically, we asked Mturk annotators to identify the topic of each presented generic summary. This process helped us later to facilitate and quickly identify irrelevant aspects in the curation phase in Section \ref{sec:curation-phase}. 

\subsection{Dataset Examples}\label{appendix:dataset-examples}
Table \ref{tab:dataset-detailed-examples} shows the aspects labels and the corresponding aspect-based summaries, which belong to 3 randomly selected different document sets in our OpenAsp.
Each document set is represented by its topic name.

\subsection{Details on OABS datasets}\label{appendix:open_related_datasets}
Table \ref{tab:open-datasets} presents more details regarding the OABS datasets  from \autoref{tab:related-datasets}.

\begin{table*}[t]
\resizebox{\linewidth}{!}{%
\begin{tabular}{c|c|c|c|c|c}
\toprule
\textbf{Dataset} & \textbf{\begin{tabular}[c]{@{}c@{}}Avg \# Tokens\\in Input\\Doc/Docset\end{tabular}} & \textbf{\begin{tabular}[c]{@{}c@{}}Avg \# Tokens\\in Output\\Summary\end{tabular}} & \textbf{\begin{tabular}[c]{@{}c@{}}Avg \# Tokens\\in Input\\Aspect Label\end{tabular}} & \textbf{\begin{tabular}[c]{@{}c@{}}Avg \# Aspects\\per Topic\end{tabular}} & \textbf{Aspect Label Examples}                                                                                                                                                                                                                                                     \\ \midrule
\begin{tabular}[c]{@{}c@{}}AnyAspect\\\citep{tan2020summarizing}\end{tabular}        & \begin{tabular}[c]{@{}c@{}} 818\end{tabular}           & \begin{tabular}[c]{@{}c@{}} 23\end{tabular}            & \begin{tabular}[c]{@{}c@{}} 1.6\end{tabular}                      & 7.0                                                             & \begin{tabular}[c]{@{}c@{}}kentucky, \\ gray, \\ mr tizzard, \\ julie bishop, \\ bernabeu, \\ watson, \\ kola, \\ cash, \\ cnn, \\ australia\end{tabular}                                                                                                                          \\ \midrule
\begin{tabular}[c]{@{}c@{}}OASUM\\ \citep{yang2022oasum}\end{tabular}             & \begin{tabular}[c]{@{}c@{}}1615\end{tabular}            & \begin{tabular}[c]{@{}c@{}}40\end{tabular}     & \begin{tabular}[c]{@{}c@{}}2.7\end{tabular}                      & 1.8                                                             & \begin{tabular}[c]{@{}c@{}}History, \\ Career, \\ Background, \\ Geography, \\ Life, \\ Reception, \\ Description, \\ Early life, \\ Demographics, \\ Production\end{tabular}                                                                                                       \\ \midrule
\begin{tabular}[c]{@{}c@{}}\textbf{\openasp{}}\\\textbf{(Ours)}\end{tabular} & \begin{tabular}[c]{@{}c@{}}7,930 \end{tabular}         & \begin{tabular}[c]{@{}c@{}} 96\end{tabular}            & \begin{tabular}[c]{@{}c@{}}3.6 \end{tabular}                      & 3.1                                                             & \begin{tabular}[c]{@{}c@{}}Hurricane impact, \\ Drug treatments, \\ Custody Ruling, \\ British response, \\ Elian's rescue, \\ Availability of gas, \\ Removal of the guardhouse, \\ Opinions on Hillary Clinton, \\ Expenses controversy, \\ Alterations of waterways\end{tabular} \\ \bottomrule
\end{tabular}
}
\caption{Statistics on the OABS datasets from \autoref{tab:related-datasets}. In the first two datasets, the input topic is a single document, while in our dataset it is a document \textit{set}. The example aspect labels in the right-most column are taken from sampled topics from across the respective dataset.}
\label{tab:open-datasets}
\end{table*}
\begin{table*}[ht]
\centering
\small
\resizebox{\textwidth}{!}{%
\begin{tabular}{l|l}
\hline
\textbf{Model}          & \textbf{Summary Text}                                                                                                                                                                                                                                                                                                                                                                                                                                                                                                                                                                                                                                                                                                                                                                                                                  \\ \hline
\BARTSumm 1k S-T5 & \begin{tabular}[c]{@{}l@{}}The launch was scrubbed four minutes before liftoff because of bad weather.  \\ 
It was rescheduled for 8:31 a.m. EDT on March 8, 1989.  \\ 
The countdown began Saturday afternoon, April 27, 1989, and  the countdown for the \\ launch of the new shuttle, Discovery, began at 3:37pm.  \\ 
Five astronauts climbed into the crew cabin just as dawn broke over the Kennedy Space Center.  \\ 
They were ready to go out of the cabin at any time to assist with repairs.  \\ 
Mission Control ordered space-walking astronauts to be ready to help with the shuttle's final\end{tabular}                                                                                                                                                                                                                             \\ \hline
\PRIMERASumm 1K S-T5 & \begin{tabular}[c]{@{}l@{}}The countdown began Saturday afternoon for a launch set for 8:31am EDT Tuesday.\\
The launch was scrubbed four minutes before liftoff.\end{tabular}                                                                                                                                                                                                                                                                          \\ \hline
\PRIMERASumm 4K S-T5 & \begin{tabular}[c]{@{}l@{}}The launch was delayed for two weeks after a faulty power unit forced the National Aeronautics \\ and Space Administration to halt the first launch four minutes before liftoff on April 10, 1989.\\
The second attempt to launch the Hubble Space Telescope was successfully launched on June 8, 1990.
\end{tabular}                                                                                                                                                                                                                                                                                                                                                                                                                                              \\ \hline
\ChatGPTSumm 4K S-T5    & \begin{tabular}[c]{@{}l@{}}The Hubble Space Telescope was launched into orbit by the Discovery shuttle, \\ marking a significant moment for NASA. \\ The telescope, which cost \$1.5 billion, was released from the shuttle's mechanical arm and \\ is expected to provide significant scientific data by satellite in a month or two. \\ The launch was delayed by two weeks due to a faulty power unit, but the procedure went smoothly \\ and the telescope was released successfully. \\ The orbit was dictated by the need to put the telescope above the Earth's distorting atmosphere. \\ The mission was considered to be the most important shuttle mission to date.\end{tabular}                                                                                                                                                                      \\ \hline
\ChatGPTStages 4K       & \begin{tabular}[c]{@{}l@{}}Aspect Based Summary:\\\\The Hubble Space Telescope was successfully launched into orbit by the Discovery \\ shuttle after several delays. \\ The telescope was deployed with the help of the shuttle's mechanical arm and two solar panels \\ that power its instruments. \\ It had some problems with its high-speed antennas and its science instruments were on hold, \\ but it is expected to provide significant scientific data by satellite in a month or two. \\ The telescope is considered the most important shuttle mission to date and is expected to provide \\ clues to the age and size of the universe, and it may solve the mysteries of quasars, pulsars, and \\ black holes and even find stars with planets that could conceivably support life.\end{tabular}                                                                                                                                                                                                        \\ \hline
\ChatGPTStages 16K      & \begin{tabular}[c]{@{}l@{}}The Hubble Space Telescope was successfully launched into orbit by the space shuttle Discovery. \\ The launch was delayed multiple times due to technical problems and the Challenger accident. \\ The telescope, named after astronomer Edwin P. Hubble, will allow astronomers to study stars \\ and galaxies that are billions of years old and provide insights into the origins and \\ expansion of the universe. \\ The telescope's mission is expected to last 15 years and it is equipped with advanced \\ instruments to capture high-resolution images and collect scientific data.\end{tabular} \\ \hline
\textbf{Gold}           & \textbf{\begin{tabular}[c]{@{}l@{}}Deployment of the Hubble Space Telescope (HST), originally scheduled for 1983,  \\ was delayed by technical problems and the Challenger accident in 1986 and did not \\ take place until April 1990. \\ A successful launch of space shuttle Discovery with the HST aboard took place at \\ 8:34 a.m. April 24, 1990 and Discovery ascended into orbit 380 miles above Earth. \\  On April 25 the HST was placed in orbit on its own with Discovery trailing behind \\ to provide astronaut assistance if needed. \\ The HST's aluminum aperture door was successfully opened on April 27 exposing \\ the telescope's 94.5- inch mirror to starlight.  \\ By April 28 there were still problems with the electronic and communications systems, but \\ procedures were in place to take care of these, the HST was in orbit, \\ and the Discovery astronauts prepared for return to earth on April 29, 1990.\end{tabular}}                                                                                                                                                                                                                                             \\ \hline
\end{tabular}

}

\caption{The generated aspect-based summaries that refer to the aspect \emph{'Launch into orbit'} of the topic \emph{'Hubble Space Telescope'}. The bottom line contains the Gold summary in \textbf{Bold}.}

\label{tab:orbit}
\end{table*}
\begin{table*}[ht]
\centering
\small
\resizebox{\textwidth}{!}{%
\begin{tabular}{l|l}
\hline
\textbf{Model}          & \textbf{Summary Text}                                                                                                                                                                                                                                                                                                                                                                                                                                                                                                                                                                                                                                                                                                                                                                                                                  \\ \hline
\BARTSumm 1k S-T5 & \begin{tabular}[c]{@{}l@{}}Unemployment in France is one of the country's main economic problems. \\ Unemployment is a major problem because of a combination of factors including \\ lack of development of job market and lack of solidarity among people. \\ The French Observatory of Economic Circumstances expects an unemployed\\ population of 140,000 by the end of the year and 120,000 in the following year. \\ France is running a near-record 12.4\% unemployment rate with a jobless population \\ of  3.1 million, one-third of whom has been unemployed for more than a year.\end{tabular}                                                                                                                                                                                                                             \\ \hline
\PRIMERASumm 4K S-T5 & \begin{tabular}[c]{@{}l@{}}Unemployment in France has long been a major problem, with a 12.4\% unemployment rate. \\ France's unemployment rate dropped slightly to 11.6\% in October from 11.7\% the previous \\ October, but is still high and has not reversed over the past 15 years. \\ Unemployment is fueled by a combination of factors, including a weak economy, a lack of \\ development of the job market, and social exclusion. \\ Also, there is concern over the high cost of unemployment insurance and unemployment \\ compensation.\end{tabular}                                                                                                                                                                                                                                                                        \\ \hline
\PRIMERASumm 1K S-T5 & \begin{tabular}[c]{@{}l@{}}Unemployment in Europe is one of the most important problems facing the country. \\ The unemployment rate in France is 11.2\%, up from 11.7\% since the Second World War. \\ In 1982 unemployment peaked at 9.7\%, the first full year of Reagan's administration, \\ while in 1975 unemployment in Germany more than doubled to 3.4\%.\end{tabular}                                                                                                                                                                                                                                                                                                                                                                                                                                                            \\ \hline
\ChatGPTSumm 4K S-T5    & \begin{tabular}[c]{@{}l@{}}France has been struggling with high unemployment rates, which have been a major economic \\ hurdle for the country. \\ The unemployment rate in France stands at 12.5\%, with 3.1 million people unemployed, \\ one-third of which have been unemployed for more than a year. The major causes of \\ unemployment in France are attributed to the economic crisis, lack of job opportunities, \\ high labor costs, and structural impediments in the functioning of labor markets. \\ The rising crime rate among youths from working-class suburbs that have been hard-hit by \\ unemployment is also a growing concern.\end{tabular}                                                                                                                                                                      \\ \hline
\ChatGPTStages 4K       & \begin{tabular}[c]{@{}l@{}}High unemployment rates in France are attributed to various factors such as high labor costs, \\ low economic growth, lack of job opportunities, insufficient social allowances, and the lack \\ of development of the job market. \\ The government has taken measures such as increasing unemployment benefits, \\ providing financial aid to the jobless, and subsidizing employment programs. \\ The OECD recommends raising wage flexibility, reforming the minimum wage system, and \\ reducing the costs of the social protection system to address the problem of unemployment.\end{tabular}                                                                                                                                                                                                        \\ \hline
\ChatGPTStages 16K      & \begin{tabular}[c]{@{}l@{}}The high unemployment rates in France are attributed to various factors, including economic \\ crisis, lack of job opportunities, rigid labor market regulations, social exclusion, and a rising \\ crime rate among youths from working class suburbs. \\ The government's efforts to address the issue through measures such as increasing social \\ allowances and unemployment benefits have been met with protests and demands for \\ further action. \\ The need for wage flexibility, reforming the minimum wage system, and reducing restrictions \\ on mass layoffs have also been highlighted as potential solutions. \\ Additionally, concerns have been raised about the high costs of France's social protection \\ system and the need to make the benefit system less generous.\end{tabular} \\ \hline
\textbf{Gold}           & \textbf{\begin{tabular}[c]{@{}l@{}}Critics attributed France's unemployment to its near-zero growth economic \\ environment, to its focus on fairness in income distribution, to rigidity in budget policy, \\ to declines in social policy, and to layoffs in the public sector. \\ Additional unemployment causes cited were high costs for low-skilled workers, \\ high unemployment insurance and compensation costs, and over-regulation of industry. \\ Critics suggested using the money paying for social aid programs to be used instead \\ for investment in jobs.\end{tabular}}                                                                                                                                                                                                                                             \\ \hline
\end{tabular}

}

\caption{The generated aspect-based summaries that refer to the aspect \emph{'Reasons for high unemployment rates'} of the topic \emph{'Unemployment in France'}. The bottom line contains the Gold summary in \textbf{Bold}.}

\label{tab:reason_unemployment}
\end{table*}
\begin{table*}[ht]
\centering
\small
\resizebox{\textwidth}{!}{%
\begin{tabular}{c|p{0.13\linewidth}p{0.8\linewidth}}
\textbf{Topic} & \textbf{Aspect Label} & \textbf{Aspect-based Summary} \\ \hline
\multirow{3}{*}{\rotatebox[origin=c]{90}{Senator Dianne Feinstein~~~~~~~~~~~~~~~~~~~~~~~~~}}     & 2000 Senate Campaign                      & She started raising \$15 million for a second term in the Senate.Her 2000 Senate campaign blamed Republicans for rejecting her patients' rights bill, tobacco regulation, and gun purchase restrictions.She ran on improving health care, preserving Lake Tahoe, restricting gun sales, eliminating the gas additive MTBE, and bringing together opponents in California's water wars.Opponent Representative Tom Campbell (R-CA), a Stanford Law professor and pro-abortion rights, made her Senate race nationally significant and her toughest since 1990, when she was booed at a state Democratic convention for supporting the death penalty.                                                                                                                                                              \\ \cline{2-3}
                                                                        & Political Positions                       & In 1999 she authored a failed resolution censuring President Clinton, but voted against both articles of impeachment.Her 2000 Senate campaign blamed Republicans for rejecting her patients' rights bill, tobacco regulation, and gun purchase restrictions.She ran on improving health care, preserving Lake Tahoe, restricting gun sales, eliminating the gas additive MTBE, and bringing together opponents in California's water wars.A centrist, her support cuts across lines of party, ethnicity and gender.She supported deficit reduction.                                                                                                                                                                                                                                                              \\  \cline{2-3}
                                                                        & Achievements of Feinstein's Senate Career & By 1998 Senator Diane Feinstein (D-CA) was the senior Democrat on the Judiciary Committee's panel on technology, terrorism and government information.Her 1984 ban on assault weapons and a Mojave Desert national park were crowning achievements of her Senate career.Feinstein amended a trade bill to eliminate sanctions against Sub-Saharan African countries producing copycat US AIDS drugs.In 2000 Feinstein introduced legislation requiring gun owners to obtain licenses and register.Feinstein successfully rebutted opponent's erroneous charges that her support for normalizing China trade relations benefited her husband and that she had hidden related financial disclosure information.Senators Feinstein and Boxer were the first two women co-chairs of a national political convention. \\ \hline
\multirow{3}{*}{\rotatebox[origin=c]{90}{William Clinton~~~~~~~~~~~~~~~~~~~~~}}              & Scandals                                  & His college career was threatened by the Vietnam War, but he arranged deferments until induction seemed unlikely, then drew a high lottery number, avoiding military service.This incident drew criticism during subsequent political campaigns.He and his wife, Hillary, invested in a real estate venture, Whitewater, and the related failure of a savings and loan.The scandal simmered for years.In the presidential campaign of 1992 he defended his record in Arkansas and his personal and draft history.                                                                                                                                                                                                                                                                                                \\  \cline{2-3}
                                                                        & Early life                                & William Clinton showed intelligence and promise from childhood.An overachiever at school, his home life was punctuated with long discussions on a variety of subjects including desegregation and social justice.After graduation he went on to Georgetown, Oxford (as a Rhodes Scholar) and Yale Law School.His college career was threatened by the Vietnam War, but he arranged deferments until induction seemed unlikely, then drew a high lottery number, avoiding military service.                                                                                                                                                                                                                                                                                                                       \\  \cline{2-3}
                                                                        & Political career                          & Returning home, he ran successfully for governor in 1978 at age 32 but was defeated for reelection.Older and wiser, he ran successfully three more times, making improvements in education, the economy and welfare.In the presidential campaign of 1992 he defended his record in Arkansas and his personal and draft history.A German editorial summarizes Clinton's early presidency.Despite some blunders, a year later he enjoys a 60\% popularity rating; the economy is up, prices stable and interest and the deficit are down.                                                                                                                                                                                                                                                                          \\ \hline
\multirow{4}{*}{\rotatebox[origin=c]{90}{{\scriptsize The tumultuous making of `Gone with the wind'~~~~~~~}}} & Problems with the cast                    & Clark Gable was under contract to Selznick's father-in-law, who finally ”loaned” him to Selznick, with plenty of strings attached—andGable wasn't thrilled about it.Original director George Cukor didn't get along with Gable and was eventually replaced by Fleming, who didn't get along with Vivien Leigh.”Leigh hated Fleming.With a passion.Fleming hated her.Clark Gable hated David … Everybody hated David,” an assistant said.Fleming quit before returning.                                                                                                                                                                                                                                                                                                                                           \\  \cline{2-3}
                                                                        & David Selznick reluctant to make the film & In Entertainment Weekly, Chris Nashawaty tells the story of the film's making, which centered on producer David Selznick.He was at first reluctant to make the film, despite a glowing review of the book by one of his employees.”I am absolutely off my nut about this book,” Katharine Brown wrote, finally convincing him to take action.                                                                                                                                                                                                                                                                                                                                                                                                                                                                    \\  \cline{2-3}
                                                                        & The premiere sparked racial tension       & And the trouble was far from over: Racism plagued the various premieres, with black cast members in many cases banned from attending, the Los Angeles Times reports.That prompted anger from Selznick, the AP reports; Gable, meanwhile, had already stood against segregated toilets on set, threatening to bail on the film, according to a Life magazine book cited by the Times.                                                                                                                                                                                                                                                                                                                                                                                                                             \\  \cline{2-3}
                                                                        & Script writing challanges                 & Among the challenges: The first writer dropped out after spending months on the script.After a number of other writers tried their hand, including Selznick himself, writer Ben Hecht took it on, but there was no time for him to read Margaret Mitchell's book.So Selznick and director Victor Fleming ”stayed up all night acting out the story for him.”                                                                                                                                                                                                                                                                                                                                                                                                                                                     \\ \hline
\end{tabular}%
}
\caption{The aspect labels and corresponding aspect-based summaries that belong to 3 random selected document sets from our \openasp{} dataset. A topic name is assigned to each document set.}
\label{tab:dataset-detailed-examples}
\end{table*}

\end{document}